\documentclass[11pt]{article}
\usepackage[margin=1in]{geometry}
\usepackage{graphicx}
\usepackage{tabularx}
\usepackage{array}
\usepackage[table]{xcolor}
\usepackage{booktabs}
\usepackage{amsmath,amssymb}
\usepackage[numbers,sort&compress]{natbib}
\usepackage{caption}
\usepackage{subcaption}
\usepackage{xcolor}
\usepackage{url} 
\usepackage{authblk} 
\usepackage{xcolor}
\usepackage{pifont}
\usepackage{makecell}
\usepackage{multirow}
\usepackage{booktabs}
\usepackage{ragged2e}
\newcommand{\red}[1]{\textcolor{black}{#1}}
\usepackage[numbers,sort&compress]{natbib}
\newcommand{\cmark}{\textcolor{green}{\ding{51}}}%
\newcommand{\xmark}{\textcolor{red}{\ding{55}}}%

\usepackage{hyperref} 
\usepackage{xurl}

\title{\textbf{TeachLM: Post-Training LLMs for Education Using Authentic Learning Data}}

\author[1]{\bf Janos Perczel\thanks{janos@polygence.org (Corresponding author)}}
\author[1]{\bf Jin Chow\thanks{jin@polygence.org}}
\author[2]{\bf Dorottya Demszky\thanks{ddemszky@stanford.edu}}

\affil[1]{Polygence}
\affil[2]{Stanford University}
\date{\today}

\begin{document}
\maketitle

\begin{abstract}
The promise of generative AI to revolutionize education is constrained by the pedagogical limits of large language models (LLMs). A major issue is the lack of access to high-quality training data that reflect the learning of actual students. Prompt engineering has emerged as a stopgap, but the ability of prompts to encode complex pedagogical strategies in rule-based natural language is inherently limited. To address this gap we introduce TeachLM -- an LLM optimized for teaching through parameter-efficient fine-tuning of state-of-the-art models. TeachLM is trained on a dataset comprised of 100,000 hours of one-on-one, longitudinal student-tutor interactions maintained by Polygence, which underwent a rigorous anonymization process to protect privacy. We use parameter-efficient fine-tuning to develop an authentic student model that enables the generation of high-fidelity synthetic student–tutor dialogues. Building on this capability, we propose a novel multi-turn evaluation protocol that leverages synthetic dialogue generation to provide fast, scalable, and reproducible assessments of the dialogical capabilities of LLMs. Our evaluations demonstrate that fine-tuning on authentic learning data significantly improves conversational and pedagogical performance -- doubling student talk time, improving questioning style, increasing dialogue turns by 50\%, and greater personalization of instruction.
\end{abstract}

\section{Introduction}



In his seminal 1984 study, educational psychologist Benjamin Bloom demonstrated that one-on-one tutoring can yield learning gains two standard deviations above those achieved through traditional classroom instruction~\cite{bloom1984twosigma}. Given the high cost of personalized tutoring, the advent of generative AI has raised hopes of scaling effective one-on-one learning to students worldwide~\cite{worldbank2022learningpoverty,kestin2024tutoring,wang2024tutorcopilot,desimone2025chalkboards,henkel2024ghana}. Yet despite the rapid adoption of AI tools such as ChatGPT, Gemini, and Claude by millions of learners~\cite{openai2025collegechatgpt,civicscience2025aifrontier}, these technologies have so far failed to realize that promise~\cite{zhai2024overreliance}. For instance, a recent University of Pennsylvania study found that unfettered access to GPT-4 for math tutoring can harm educational outcomes~\cite{bastani2025guardrails}. Similarly, an MIT study reported that participants who used LLMs exhibited significantly reduced brain connectivity and struggled to quote from answers they wrote just minutes earlier~\cite{kosmyna2025cognitivedebt}.

\begin{figure}[h]
    \centering

    \begin{minipage}[c]{0.67\textwidth}
        \centering
        \includegraphics[width=\linewidth]{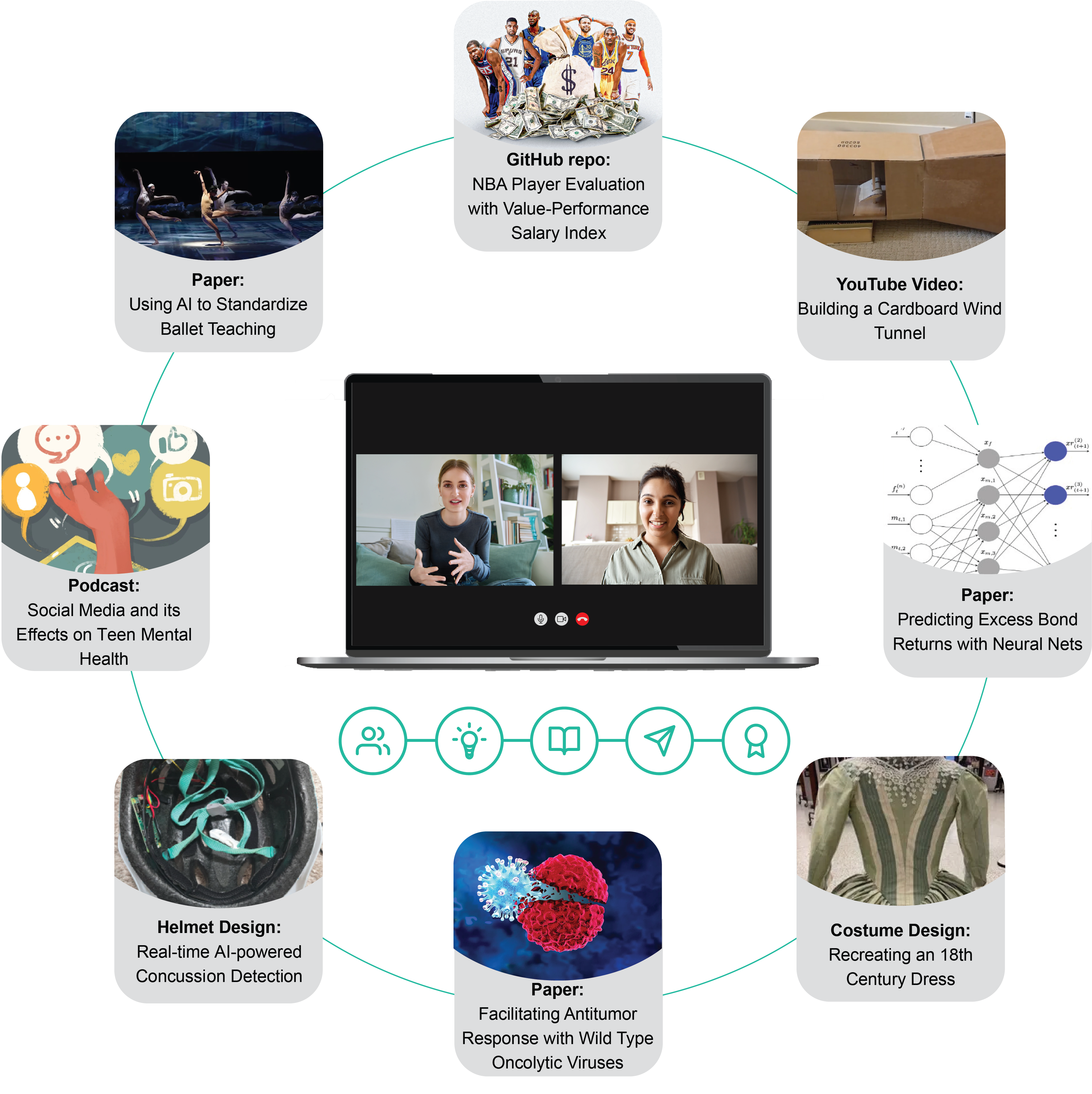}
    \end{minipage}%
    \hfill
    \begin{minipage}[c]{0.32\textwidth}
        \centering
        \includegraphics[width=\linewidth]{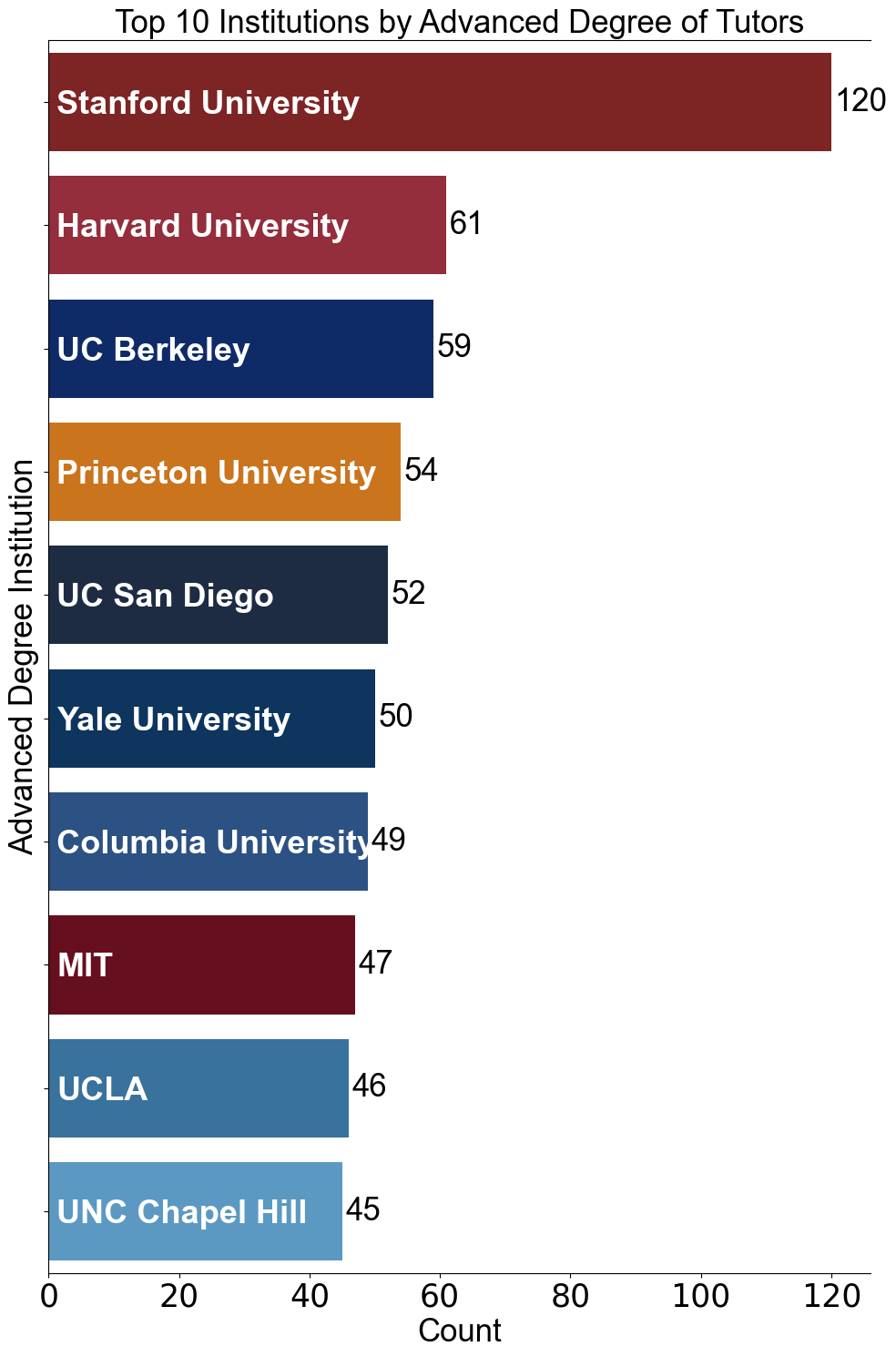}
    \end{minipage}

    \caption{Left: Illustration of the Polygence program and project outcomes. Students meet online with tutors pursuing or holding advanced degrees (PhD, MD, JD, MBA, etc.) to take projects from ideation, to exection, to showcasing. Project outcomes range from academic papers to creating podcasts to engineering physical devices. Topics range from AI to cancer biology to sport analytics. Right: Top 10 institutions represented by the advanced degrees pursued or held by Polygence tutors.}
    \label{fig:polygence_and_institutions}
\end{figure}

A fundamental issue with LLMs is that they have been optimized to act as ``helpful assistants'' that maximize productivity and minimize cognitive labor~\cite{wei2021flanzeroshot, peng2023instructiontuninggpt4, bai2022helpfulharmless, ouyang2022instructgpt,gemini2023multimodal, anthropic2024claude3, meta2024llama3, achiam2023gpt4, jiang2023mistral7b}. This contrasts with the natural friction that expert teachers introduce into learning (for example, by withholding the right answer and prompting students to first attempt a response)~\cite{bjork2020desirabledifficulties,lodge2018confusionreview, bjork2011desirabledifficulties}. Effective tutoring also requires dynamic adaptation to learners' states of mind as opposed to one-size-fits-all instructional designs~\cite{duboulay2018motivationalITS}. This tendency toward friction minimization and sycophantic behavior—prioritizing compliance over pedagogy—is systematically encoded in model parameters through supervised fine-tuning~\cite{wei2021flanzeroshot} and reinforcement learning from human feedback (RLHF)~\cite{christiano2017deep, ziegler2019finetuning}. These processes rely on datasets produced by human annotators instructed to provide responses that maximize completeness while minimizing the number of conversational turns~\cite{ouyang2022instructgpt}.

Off-the-shelf LLMs can be steered towards improved pedagogy to a limited extent through prompt engineering~\cite{stanford2024promptengineering,anthropic2024claudeedu,openai2024studymode,google2024guided,modi2025evaluating,wang2025chatgptmeta}, but prompting alone cannot resolve the underlying issues. No finite set of rules or instructions, however sophisticated, can capture the complexity and nuance of high-quality human pedagogy, which necessarily adapts to diverse learners, study contexts, and student goals~\cite{learnlm2024}. We encountered these limitations first-hand in our attempts to build a project-based tutor---called Polypilot---with GPT-4, where iterative prompt refinement led to an endless cycle of increasingly elaborate instructions in response to new scenarios and edge cases. Such pedagogical constraints persist even in the most recent education-focused LLMs, including Anthropic’s Learning Mode~\cite{anthropic2024claudeedu}, OpenAI’s Study Mode~\cite{openai2024studymode}, and Google’s Guided Learning~\cite{google2024guided} (integrating LearnLM~\cite{modi2025evaluating}). For example, when confronted with student confusion, these models typically default to rephrasing the problem rather than diagnosing its underlying source. Similar challenges arise in prompt-engineered student simulators, which typically lack the authenticity and diversity needed to represent the full spectrum of learning personas~\cite{markel2023gpteach}.

Post-training frontier models on domain-specific data has recently led to rapid progress toward human-level performance across a range of domains, including coding, law, and science~\cite{wang2025aethercode, deepmind2025geminiICPC, Xie2023DARWIN, hu2025breakingbarriers}. This progress has been made possible by the availability of copious amounts of high-quality training data generated by human annotators who adhere to clearly defined standards of excellence~\cite{kopf2023openassistant}. We expect similar progress to be possible in education given sufficient availability of training data and well-defined success metrics~\cite{sutton2019bitterlesson}. Recently, Google's LearnLM team demonstrated that supervised fine-tuning of LLMs on synthetic data can improve their performance on a range of education-related benchmarks~\cite{learnlm2024}. Post-training LLMs for education is especially important given that modeling effective pedagogical behavior inherently entails {\it both} an expert teacher {\it and} an authentic learner. Without access to a realistic student model---either a high-fidelity simulator or real students, the latter being hard to scale and ethically problematic---benchmarking candidate teacher models is severely limited. 


A persistent challenge for post-training educational models is the scarcity of authentic learning data from human teachers and students due to logistical barriers, privacy protections, and concerns about data quality~\cite{vasselli2023naisteacher,macina2023mathdial,tack2022aiteachertest,hicke2023efficacy,abdelghani2023gpt,macina2023opportunities,stasaski2020cima,caines2020chatroom,suresh2022discursive,demszky2022ncte}. Moreover, human annotators cannot reliably simulate the active learning processes of students or replicate expert teacher practices without engaging with real learners, making the on-demand collection of such data particularly difficult. To address these limitations, researchers from MIT, Carnegie Mellon University, and Cornell University have launched the National Tutoring Observatory~\cite{nationaltutoringobservatory2025}, with support from the Gates Foundation, the Chan Zuckerberg Initiative~\cite{kizilcec2025national}, and the National Science Foundation~\cite{nsf2321499}. The initiative aims to collect and open-source one million teacher–student interactions to inform the development of AI tutoring tools~\cite{nationaltutoringobservatory2025}. While this represents an important step toward alleviating the critical shortage of educational training data, additional work will be required to realize the full potential of post-training models in education.

\red{In this preliminary report, we present a case study on post-training LLMs for education, drawing on data from the Polygence platform~\cite{polygence2025} consisting of over 100{,}000 hours of one-on-one, project-based tutoring sessions between PhD-level tutors and students 
across more than 150 subjects (Fig.~\ref{fig:polygence_and_institutions}). Data was curated consistent with the platform's terms of use and privacy policy, reflecting participant opt-outs, and underwent a rigorous anonymization process to protect privacy. Using this dataset, we fine-tune a high-fidelity student model to benchmark frontier LLMs on six multi-turn conversational and pedagogical evaluations. We also fine-tune a teacher model, TeachLM, and demonstrate that it significantly outperforms off-the-shelf models on these benchmarks. Our main contributions are as follows:}
\begin{enumerate}
    \item \red{We develop a pipeline for transcribing, diarizing, and cleaning single-track audio recordings to produce high-quality dialogical data for post-training.}
    \item \red{We show that student data enables the training of authentic student models, which are essential for scalable and reproducible evaluation of LLMs’ pedagogical capabilities.}
    \item We benchmark off-the-shelf LLMs against human tutors across six education-focused evaluations, highlighting systematic differences in conversational and engagement metrics.
    \item We \red{demonstrate that parameter-efficient fine-tuning of state-of-the-art LLMs on authentic learning data substantially improves their pedagogical performance.}
\end{enumerate}

We conclude by outlining the limitations of our current approach and identifying next steps for refining the post-training process and evaluating its efficacy.

\section{Background: Improving LLM Pedagogy}

Addressing the pedagogical limitations of LLMs through both post-training and prompt engineering has been an active area of research. Below we review a few of these efforts.

\subsection{LearnLM: Google's Fine-Tuned Model}
A pioneering effort by Google's LearnLM team~\cite{google2025learnlm} has focused on improving LLMs for education by post-training. Their early efforts focused on the targeted collection of human tutoring data and the generation of synthetic data under the guidance of education experts~\cite{learnlm2024}. While the on-demand collection of human contractor data from impersonating students proved too noisy for training, their synthetic data allowed the training of an LLM optimized for pedagogy, called LearnLM~\cite{learnlm2024}. LearnLM demonstrated improvements across a range of measurable benchmarks, such as guiding students to answers, promoting engagement, or identifying misconceptions~\cite{learnlm2024}. While synthetic data alone cannot fully capture the nuances of effective human pedagogy and authentic student learning, these improvements highlight the promise of using domain-specific data for post-training LLMs for education. Recently, the LearnLM Team at Google has shifted its focus from {\it post-hoc fine-tuning} of models to optimizing it for {\it pedagogical instruction following} of teacher- or developer-defined prompts~\cite{modi2024learnlm}. While acknowledging the shortcomings of prompting, they explained their decision citing the `prohibitively difficult' challenge of defining ideal AI tutoring behavior, instead leaving it to teachers and developers to decide on the desired behavior~\cite{modi2024learnlm}. They also highlighted the cost and overhead of maintaining fine-tuned models while base models are developing rapidly~\cite{modi2024learnlm}. 

\subsection{PolyPilot: Polygence's Prompt-Engineered Tutor}
\label{section:polypilot}
We first experienced the limitations of prompt engineering through a high-conviction internal product experiment in early 2024, when a dedicated engineering and product team at Polygence~\cite{polygence2025} built Polypilot ---- a dynamically prompt-engineered project-based tutor using GPT-4. PolyPilot was deployed in production after many months of user feedback and iterative development, and we collected detailed feedback from more than 70 engaged users (see Appendix~\ref{app:polypilot}). 

Given that Polygence specializes in high-quality project-based tutoring (the success of which can be judged by students' ability to deliver showcaseable artifacts), we quickly recognized the limitations of our prompt-engineered tutor and its ability to steer students towards a project outcome. We invested several months of deliberate effort to iteratively refine stage-dependent prompts and we combined the LLM with sophisticated user interfaces and constantly gathered user feedback to guide our experimentation. However, we soon reached the conclusion that the gap between a human tutor and GPT-4 was simply too large to be closed by prompt engineering. Even relatively simple tasks, such as varying the number and placement of questions or avoiding “wall-of-text” responses to better mirror human tutors, proved inconsistent with prompting. We continued our experimentation with GPT-4o, o1, Gemini 1.5, and Gemini 2.0 Flash, and progressively introduced complex RAG-based approaches to provide LLMs with high-quality examples. However, by early 2025 we concluded that, despite the rapid improvement of LLMs, prompt engineering of off-the-shelf models continues to be insufficient to build a product that can meet the standard of human tutoring.

\subsection{Anthropic's Learning Mode, OpenAI's Study Mode, and Google's Guided Learning}
\label{section:study_modes}

In recent months, large model developers have released educational LLMs that have been prompt-engineered to improve their pedagogical behavior. Examples include Anthropic's  Claude Sonnet 4 with \emph{Learning Mode}~\cite{anthropic2024claudeedu}, OpenAI's GPT 5 with \emph{Study Mode}~\cite{openai2024studymode}, and Google's Gemini 2.5 Pro with \emph{Guided Learning}~\cite{google2024guided} (which integrates LearnLM~\cite{modi2025evaluating}). None of these customized LLMs are available through application programming interfaces (APIs), making rigorous, large-scale, and repeatable evaluation of these models difficult. Nonetheless, to get a directional sense of their behavior, we manually tested all three LLMs via multi-turn conversations dozens of times with different learning scenarios (see Appendix~\ref{app:study_modes} for more details). 

Based on our experimentation, our informal assessment of these educational modes is that their pedagogical capabilities remain rudimentary, and are noticeably constrained by the limitations of rule-based prompt engineering. For example, we found that these models often miss students' learning context, default to multiple-choice-style questioning rather than asking open questions, give away answers, fail to appropriately address confusion, and struggle with verbosity. We also found that models from different providers exhibit remarkably similar behavioral patterns -- highlighting both the limited power of prompt engineering to overwrite the pedagogy-agnostic principles learned from large-scale training data and the likely consequence of major AI labs sourcing data from the same vendors (e.g., Scale AI, Surge AI, and Mercor~\cite{reuters2025surgeai_capraise}). 

\begin{figure}[h]
    \centering
    \includegraphics[width=1\linewidth]{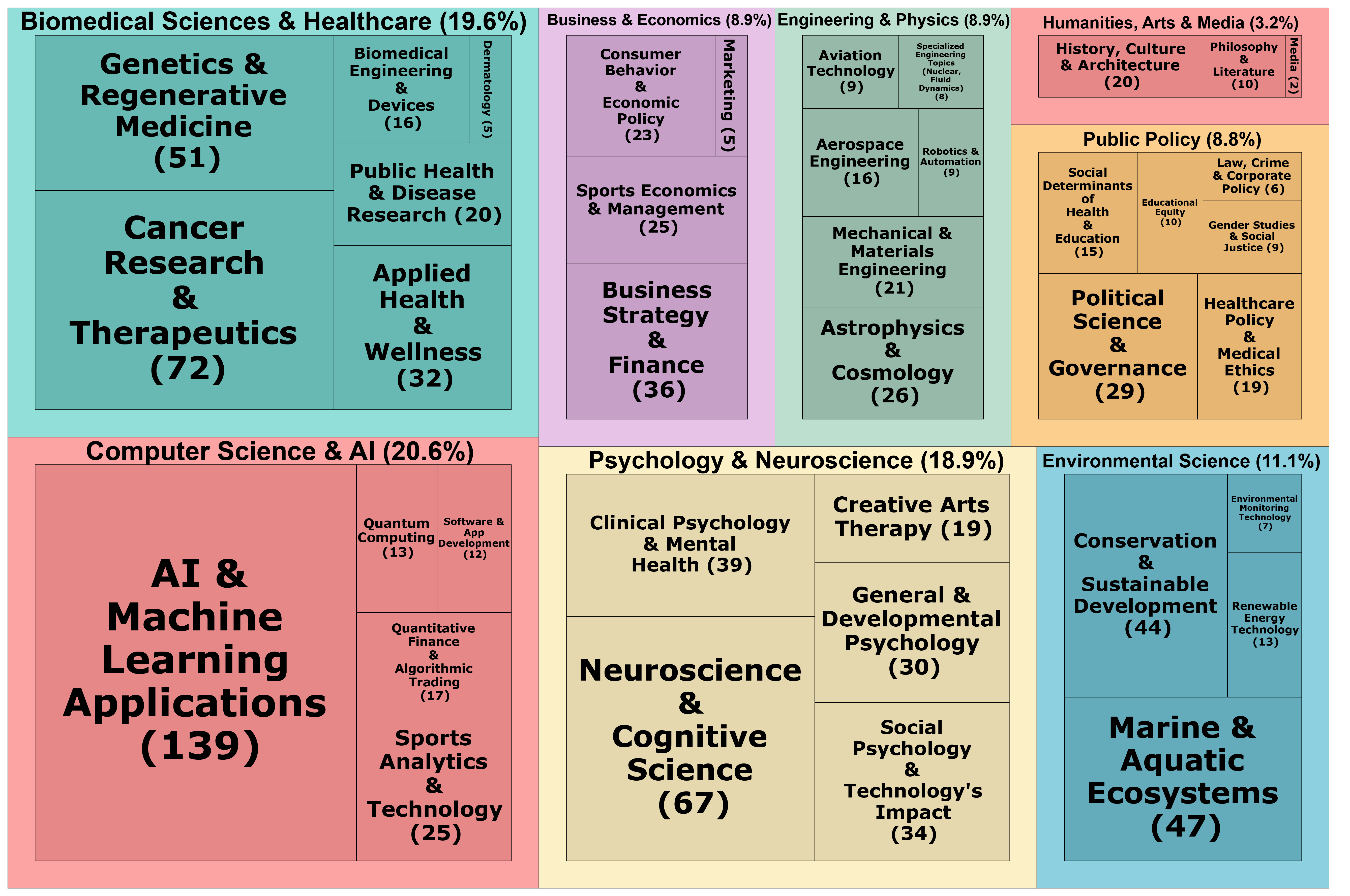}
    \caption{A squarified hierarchical map of the  distribution of project topics based on a random sample of $n=1,000$ Polygence projects. To create this map, we used a customized version of K-means clustering of project topics based on Anthropic's Clio framework \cite{tamkin2024clio} and the open-source Kura library \cite{567labs2025kura}. The size of each box is proportional to the relative frequency of each topic or topic cluster.}
    \label{fig:subjects}
\end{figure}

\section{Curating Authentic Learning Data for Post-training}
\label{section:polygence_data}

\red{After experiencing the limitations of prompt engineering while building PolyPilot, our focus shifted toward post-training LLMs on authentic learning data from Polygence.}

\subsection{Authentic Learning Dataset}
\red{The Polygence dataset comprises more than 100,000 hours of one-on-one, longitudinal student–tutor interactions in multiple modalities. This data was collected through the Polygence platform~\cite{polygence2025} consistent with the platform's terms and conditions while honoring opt-outs (see Section~\ref{section:data_privacy_security}) and underwent a rigorous anonymization process to protect privacy. The Polygence platform supports an online program for one-on-one project-based learning under the guidance of expert, PhD-level US-based mentors. Students join Polygence to engage in projects covering a wide array of subjects both in STEM and in the Humanities. Illustrative projects include building a cardboard wind tunnel \cite{polygence2020windtunnel}, sewing a historically accurate 19th-century dress \cite{olivieri2023fashionhistory}, writing a paper on using AI to standardize ballet teaching \cite{priyankaAIballet}, recording a podcast about the neuroscience of dementia \cite{polygence2020tori_memorypodcast}, and creating a helmet with an AI-powered concussion detection system~\cite{sen2024concussiondetection} (see Fig.~\ref{fig:polygence_and_institutions}).} The projects are overseen by PhD-level experts from top US research institutions. The top 10 institutions represented are shown in Fig.~\ref{fig:polygence_and_institutions}.

\begin{figure}[h!]
    \centering
    \begin{subfigure}[b]{\textwidth}
    \raggedright

        \centering
        \includegraphics[width=\textwidth]{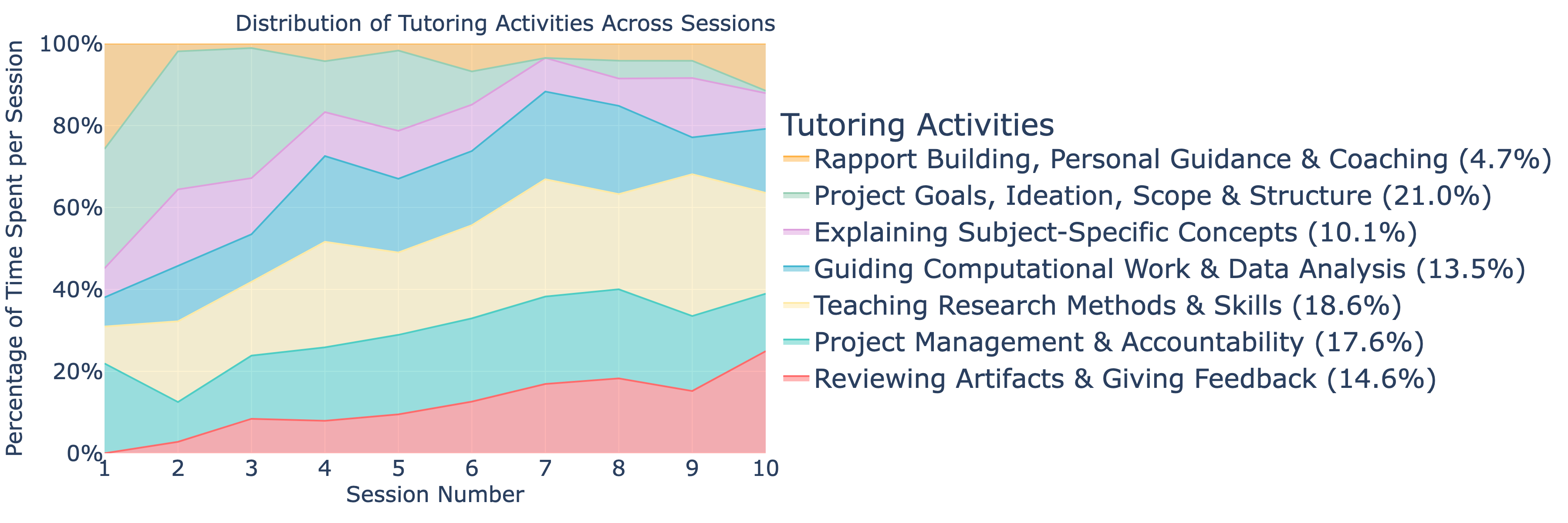}
        \label{fig:mentorship_main_only}
    \end{subfigure}

    \vspace{.0em} 

    \caption{Tutoring activity overview for $n=195$ completed 10-session projects. Each 1-hour session is segmented into 5-minute chunks, analyzed individually, and hierarchically clustered into 4 levels using using Anthropic's Clio framework \cite{tamkin2024clio} and a customized version of the open-source Kura library \cite{567labs2025kura}. The distribution of the 7 top-level tutoring categories are shown across sessions. We find a 78\% overlap with the top 10 student usage categories of ChatGPT reported by OpenAI~\cite{openai2025collegechatgpt}.}
    \label{fig:all_images_together}
\end{figure}

\begin{table}[h!]
\centering

\setlength{\tabcolsep}{4pt}
\renewcommand{\arraystretch}{1.2}
\resizebox{\textwidth}{!}{%
\begin{tabular}{
  >{\RaggedRight\arraybackslash}p{4.2cm}
  >{\RaggedRight\arraybackslash}p{4cm}
  >{\RaggedRight\arraybackslash}p{4cm}
  >{\RaggedRight\arraybackslash}p{4cm}
  >{\RaggedRight\arraybackslash}p{4cm}
  >{\RaggedRight\arraybackslash}p{4cm}
}
\toprule
\textbf{Main Category} & \textbf{Subcat 1} & \textbf{Subcat 2} & \textbf{Subcat 3} & \textbf{Subcat 4} & \textbf{Subcat 5} \\ \midrule

\cellcolor[HTML]{F2D09F} Rapport Building, Personal Guidance \& Coaching &
Building Rapport \& Personal Connection (56.8\%) &
Providing Encouragement \& Emotional Support (20.9\%) &
Advising on College \& Career Pathways (15.5\%) &
Guiding Student Reflection \& Goal Setting (6.8\%) &
 \\ \bottomrule

\cellcolor[HTML]{bdddd5}  Project Goals, Ideation, Scope \& Structure &
Structuring \& Outlining Project Deliverables (37.8\%) &
Developing \& Refining Methodology (29.6\%) &
Brainstorming \& Refining Project Ideas (17.2\%) &
Defining \& Refining Scope, Goals \& Topics (15.4\%) &
 \\ \midrule

\cellcolor[HTML]{e3c7e9} Explaining Subject-Specific Concepts &
Explaining STEM Concepts (40.2\%) &
Explaining AI/ML \& Data Science Concepts (26.3\%) &
Explaining Social Science \& Humanities Concepts (18.2\%) &
Explaining Foundational Math \& Statistical Theory (8.6\%) &
Explaining Business, Finance \& Marketing Concepts (6.7\%) \\ \midrule

\cellcolor[HTML]{95d2e4} Guiding Computational Work \& Data Analysis &
Guiding Coding \& Debugging (39.6\%) &
Guiding Data Analysis \& Interpretation (37.6\%) &
Assisting with Technical Setup \& Tools (13.6\%) &
Guiding Data Sourcing \& Preparation (9.2\%) &
 \\ \midrule

\cellcolor[HTML]{f2ebcf} Teaching Research Methods \& Skills &
Guiding Academic Writing \& Structure (48.9\%) &
Guiding Literature Review \& Source Analysis (26.6\%) &
Instruction on Citation, Formatting \& Ethics (17.2\%) &
Developing Arguments \& Integrating Evidence (5.2\%) &
Coaching Presentation \& Public Speaking Skills (2.2\%) \\ \hline

\cellcolor[HTML]{99dddd} Project Management \& Accountability &
Scheduling Sessions \& Coordinating Logistics (46.9\%) &
Managing Timelines \& Deadlines (22.6\%) &
Assigning \& Clarifying Tasks (14.3\%) &
Guiding Final Submission Process (12.4\%) &
Navigating Platform \& Admin Procedures (3.9\%) \\ \midrule

\cellcolor[HTML]{f2acb1} Reviewing Artifacts \& Giving Feedback &
Providing General Writing Feedback (42.4\%) &
Collaborative Editing \& Word-Smithing (23.9\%)&
Refining Paper \& Project Structure (16.1\%) &
Giving Feedback on Presentations \& Visuals (8.8\%) &
Reviewing Overall Project Progress (8.8\%) \\ \midrule

\end{tabular}
}
\caption{Main tutoring activities, with subcategories and percentages listed across columns obtained via hierarchical K-means clustering. \label{tab:session_activities}}
\end{table}

\red{This dataset represents a distinct opportunity to evaluate the efficacy of training LLMs on authentic learning data because of its unique characteristics:}
\begin{itemize}
    \item \red{\textbf{Longitudinal interactions}: Projects typically last 4–6 months and capture the entire learning process, including the development of student–tutor relationships over time—a key driver of instructional effectiveness \cite{dilisio2025tsr,vanherpen2024relationships,apa_relationships}.}
    \item \textbf{Full personalization}: Each project is tailored to the student’s academic needs and goals. 
    \item \red{\textbf{Multi-modal exchanges}: Each project's dataset encompasses meeting transcripts, shared documents, chat, and other modalities of student–tutor interaction.}
    \item \textbf{Outcome-oriented projects}:  Over 80\% of completed projects culminate in a showcaseable artifact, such as an academic paper, video, podcast, physical prop, or GitHub repository (see Fig.~\ref{fig:polygence_and_institutions}).
    \item \red{\textbf{Alignment with student AI usage}:  Approximately 80\% of the tutoring activities overlap with the top 10 student use cases of AI reported by OpenAI~\cite{openai2025collegechatgpt}.}
\end{itemize}

To understand the distribution of topics and activities covered in Polygence sessions, we apply K-means clustering to the session transcripts using Anthropic's Clio framework \cite{tamkin2024clio} and a customized version of the open-source Kura library \cite{567labs2025kura}. First, we focus on a high-level overview of the diverse set of topics covered by Polygence projects. We randomly sample $n=1,000$ projects and cluster them based on their project topics. Fig.~\ref{fig:subjects} shows that the majority of the topics cover computer science, biomedical topics, psychology and neuroscience. The most popular topic is AI and machine learning followed by cancer research. 

Next, we focus on the various tutoring activities that are undertaken during the live sessions. We select a representative subset of $n=195$ completed projects, which typically have 9-10 full sessions. We then divide each 1-hour session into 10–15 segments, each about 5 minutes long ($n=24,587$ chunks in total) and recursively classify them to obtain both lower-level (more specific) and higher-level categories of topics and activities. 

Fig.~\ref{fig:all_images_together} highlights how the distribution of activities between the tutor and the student shifts dynamically as the project progresses across the ten sessions. Rapport and relationship building represents up to 25\% of the first session and up to 10\% of the last session and remains an important part of the mentoring process throughout the ten sessions (e.g. providing encouragement and emotional support). Ideation and setting project goals and scope are heavily prioritized upfront, but remain 10-20\% of the time budget up to mid-way through the project -- showing the highly iterative nature of this process. Subject-specific tutoring (e.g. explaining AI concepts) and technical guidance (e.g. coding and data analysis) account for about a quarter of all time spent on the project and is evenly distributed across the entire process. Reviewing artifacts (e.g. writing and presentation feedback) gradually increases in importance as the project progresses. These details highlight the highly dynamic and iterative nature of these projects as well as the richness of longitudinal tutor-student interactions that are difficult to capture in simple, static rules or heuristics. Table~\ref{tab:session_activities} gives a detailed breakdown of each high-level activity into more specific activities.

This detailed activity map of a representative sample of tens of thousands of 5-minute session fragments allows us to map the tutoring activity on the Polygence platform to the top 27 categories of ChatGPT usage by 18-24 year old students, as reported by OpenAI \cite{openai2025collegechatgpt}. By introducing the OpenAI-defined categories into our clustering process, we find that approximately 78\% of our data covers the top 10 ChatGPT use cases by students. These include starting papers/projects, brainstorming creative projects, exploring topics, editing writing, solving mathematical problems, conducting academic research, tutoring, and drafting essays (see Appendix~\ref{app:openai_overlap} for more details).

\subsection{Data privacy, User Consent, Anonymization, and Data Security}
\label{section:data_privacy_security}

\red{Data was collected and processed consistent with our Terms of Use and Privacy Policy, and reflects participant opt-outs. In addition to the rights granted through these policies, we secure consent for every recorded session with explicit, real-time notification of participants about recording at the start of each call. Given that some tutor profiles may be publicly associated with Polygence, we took certain steps to anonymize their identities in the transcripts and remove information that could personally identify them (PII) before training, including their personal background, education, life experiences, academic expertise, personal opinions, etc.} This data anonymization was done on Polygence’s internal servers. All subsequent experimentation and model evaluation was conducted exclusively on our service provider’s enterprise-grade platforms. Those service providers are contractually required to maintain confidentiality, ensure appropriate data security protections for the data, and are prohibited from using Polygence’s data for their own AI model training.


\subsection{Transcription, Diarization, and Cleaning}
\label{section:transcript_processing}
\red{Our training pipeline is built on audio from tutoring sessions, recorded with the explicit consent of all participants as outlined in our consent framework (Section~\ref{section:data_privacy_security}). High-quality training data is a prerequisite for effective post-training. Prior work shows that quality outweighs quantity during fine-tuning (e.g., \cite{krause202510000xfidelity}). We apply a multi-step pipeline to clean and post-process audio and transcripts from tutoring sessions before text-based fine-tuning.}

\begin{figure}[h]
    \centering
    \includegraphics[width=1\linewidth]{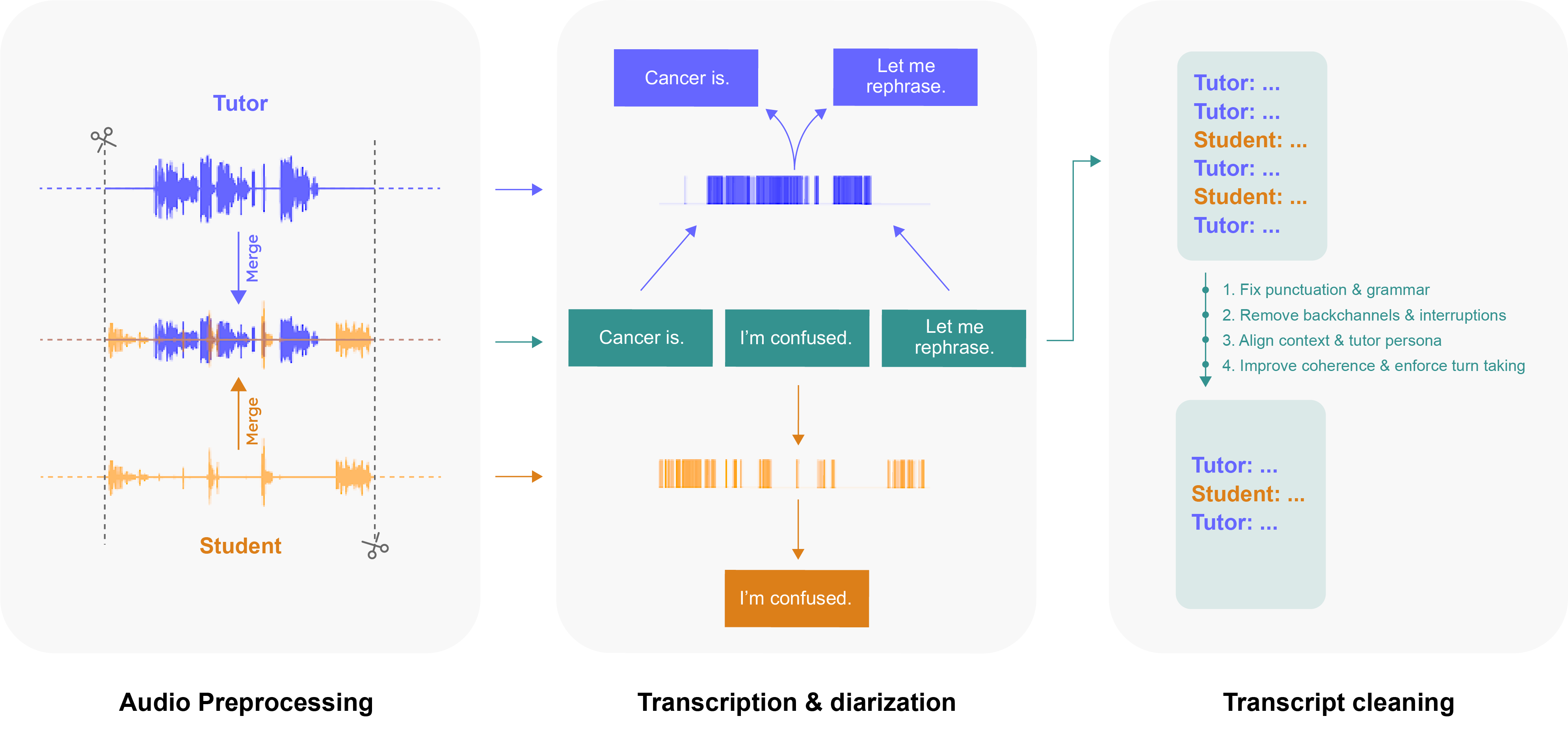}
    \caption{End-to-end transcript processing pipeline. Dual-track audio is merged and trimmed, then transcribed with high fidelity. Speaker activity masks enable accurate diarization, followed by a multi-step cleaning process (fix punctuation and grammar, remove backchannels and interruptions, align context and tutor persona, and improve coherence and enforce turn taking) to yield polished tutor–student transcripts.}

    \label{fig:transcript_processing}
\end{figure}

\paragraph{Data pre-filtering.}
\red{We retain projects with a single, uninterrupted tutor–student match. We utilize dual-track Zoom recordings (separate tracks per speaker), a feature that enables higher-fidelity transcription and reliable diarization.}

\paragraph{Audio transcription and diarization.}
As illustrated in Fig.~\ref{fig:transcript_processing}:
\begin{enumerate}
    \item \red{Assign single-speaker tracks to \textsc{Tutor} or \textsc{Student} via fuzzy matching of Zoom display names to known participants.}
    \item \red{Merge tracks into one file and trim leading/trailing silence for all files to reduce transcription hallucinations from long gaps.}
    \item \red{Transcribe the merged two-speaker audio with ElevenLabs’ API~\cite{elevenlabs2025scribe}, which yields high-fidelity text (including fillers) but unreliable speaker tags.}
    \item \red{Derive per-speaker activity masks from the single-speaker tracks.}
    \item \red{Attribute each utterance to the speaker with maximal temporal overlap; compile a diarized transcript with one speaker per statement.}
    \item Generate multiple transcript candidates and select the best using a reasoning model (Gemini 2.5 Pro), mitigating rare but consequential ASR lapses (e.g., dropped segments).
\end{enumerate}

\paragraph{Transcript cleaning.}
Using a reasoning model (Gemini 2.5 Pro), we apply:
\begin{enumerate}
    \item Normalization of punctuation and grammar.
    \item Removal of backchannels that interrupt flow (e.g., ``yeah'', ``gotcha'', ``uh-huh'').
    \item \red{Context adjustments and anonymization by removing or reframing references to human tutor identity, personal anecdotes, comments related to physical embodiment, platform names, and program-specific details.}
    \item Coherence smoothing and enforced turn-taking by merging consecutive same-speaker utterances.
\end{enumerate}

Finally, we conduct human spot checks to ensure transcripts meet the quality bar for post-training.

\section{Multi-turn Evaluations Using a Fine-Tuned Student Model}

Beyond sufficient high-quality data, effective fine-tuning also requires a clear measure of good performance. A persistent issue in evaluating LLMs pedagogical performance is the lack of industry-wide standards for multi-turn evaluations~\cite{kwan2024mteval, sirdeshmukh2025multichallenge}. This contrasts with single-turn evaluations, which have become ubiquitous in the generative AI industry, including within the educational domain~\cite{learnlm2024,modi2024learnlm, modi2025evaluating}. Here we introduce (i) a set of proxies for high-quality pedagogy in multi-turn dialogues, and (ii) a novel multi-turn evaluation protocol that combines a fine-tuned student model and a fine-tuned tutor model to generate a large number of synthetic dialogues. Each of these dialogues are then evaluated using traditional, single-shot evaluation methods and the results are then aggregated. These stochastic methods represent fast, scalable and reproducible measures of LLMs multi-turn performance.

\subsection{Defining Proxies for High-Quality Pedagogy}
\label{section:proxies}

A major challenge in education research is the lack of universally accepted pedagogical best practices~\cite{learnlm2024, dynarski2007effectiveness, li2005cognitive, klahr2013scientificrigor, ogan2023culturallyrelevant}. While certain behaviors are commonly cited as markers of effective teaching (e.g., not giving away the answer, asking questions, balancing talk time), good pedagogy is highly context-dependent, making simple heuristics insufficient to capture the full spectrum of quality instruction. Nevertheless, advancing LLM training requires well-defined evaluation criteria, and thus it is necessary to select at least \textit{some} practical proxies for high-quality pedagogy.  

In this preliminary case study, we focus on a handful of straightforward benchmarks, deferring more complex evaluations to future work. Our choice of benchmarks was informed by (i) prior academic studies (see below) and (ii) extensive user feedback from our PolyPilot experiment (see Section~\ref{section:polypilot}). Specifically, we highlight the following benchmarks:  

\begin{itemize}
    \item \textbf{Student talk-time:} The percentage of words uttered by the student relative to the total number of words in the dialogue. In a large-scale randomized controlled trial on the Polygence platform, increasing student talk time was positively associated with outcomes such as academic confidence and participant satisfaction~\cite{demszky2023liu}.  
    \item \textbf{Average number of words per tutor turn:} A measure closely tied to talk time, used to detect the prevalence of ``wall-of-text'' responses, a well-documented issue in LLM outputs~\cite{briakou2024verboseoutputs, nayab2024concisethoughts, zhang2024verbositycompensation}.  
    \item \textbf{Mean questions per interrogative turn:} PolyPilot user feedback highlighted that LLMs often display unnatural questioning styles such as asking a series of questions in the same turn. This metric captures the average number of questions per interrogative turn and serves as a proxy for human-like questioning practices, such as favoring open-ended inquiries. An interrogative turn is defined as a statement that has at least one question and questions are detected via question marks.
    \item \textbf{Number of turns before wrap-up:} LLMs are typically trained to resolve queries as quickly as possible, which can limit their ability to sustain extended and meaningful educational dialogue. This metric is defined as the number of student and tutor turns before the tutor indicates that the discussion is over (e.g. telling the student to do the assigned work and report back when they are done). 
    \item \textbf{Uncovering student background and learning context:} LLMs frequently suffer from the ``first mile problem''—jumping into explanations without first eliciting information about the student’s academic background or learning goals~\cite{cohn2025adaptive_scaffolding} (also see Appendix~\ref{app:study_modes}). To evaluate this proxy, we track 
    what percentage of all known information about a student is uncovered by a tutor during a dialogue. To do so, we extract and tabulate the information learned about the student in the original human-to-human conversation, which we treat as the theoretical maximum of 100\%. We note that even the most skilled teachers would score less than 100\% on this benchmark as every conversation is unique and  uncovers a slightly different set of facts about the student. Nevertheless, this provides a directional measure of improvement in uncovering relevant information about the student (i.e., higher scores indicate better performance).
    \item \textbf{Checking coding skills for coding projects:} A targeted version of the previous benchmark, this tests whether the model probes a student’s coding proficiency (a binary yes/no decision) before initiating technical projects. Our data indicates that nearly all human tutors begin by carefully assessing their students’ coding backgrounds and even with thoughtful calibration of project difficulty, about 20\% of coding projects still encounter challenges related to scope or student skill.
\end{itemize}

These benchmarks are intentionally simple, and in principle, any LLM could ``ace'' them through prompt-engineered rules (e.g., ``always ask about the student’s coding background before starting a computational project''). However, they serve only as \textit{proxies} for educational quality. Our central thesis is that a model with strong pedagogical capabilities would perform well on these metrics, but optimizing solely for them does not guarantee high-quality teaching.

\label{section:student_model}
\begin{figure}[h]
  \centering
  \includegraphics[width=\textwidth]{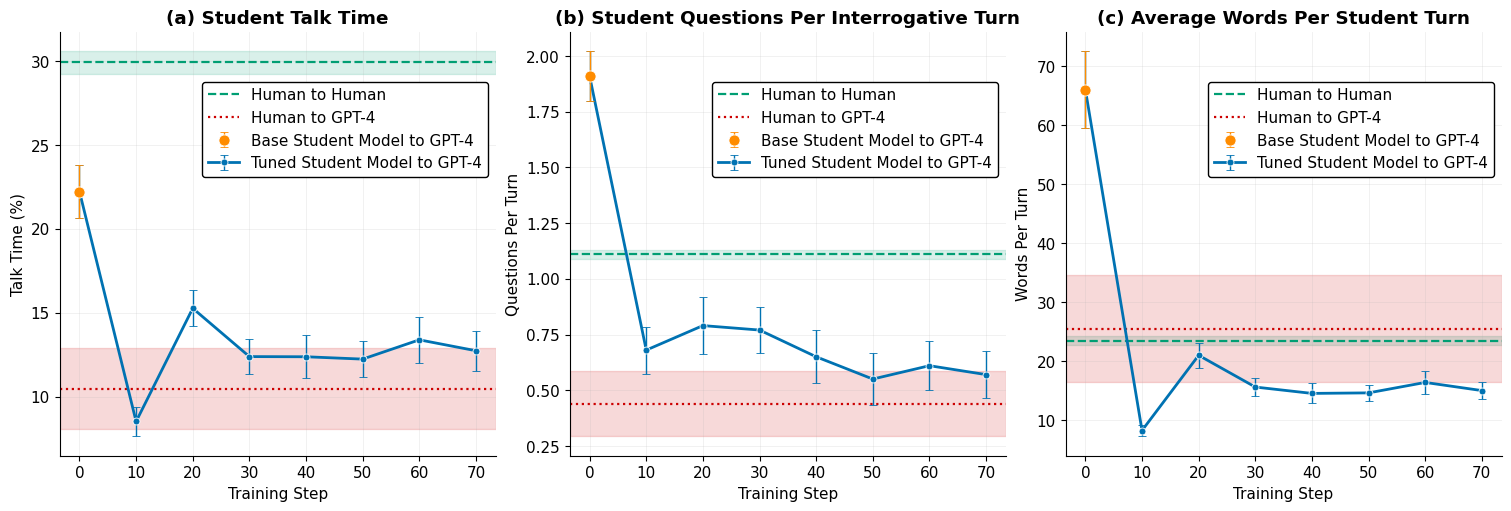}
  \caption{Comparing three core conversational statistics (talk time, questions per turn, words per turn) across four different types of dialogues: human to human, human to GPT-4, base student model (Gemini 2.0 Flash) to GPT-4, and tuned student model (Gemini 2.0 Flash tuned on Polygence student data) to GPT-4. 
  The human-to-GPT-4 conversational data was obtained from our PolyPilot experiment (Section~\ref{section:polypilot}).
  We observe that simulated conversations between two prompt-engineered base models (large orange dot) produce significantly different ($p<0.001$) conversational statistics from both dialogues involving only humans (green dashed line) and a human student and AI (red dotted line). Fine-tuning a model on student data (connected blue dots) progressively aligns its conversational statistics with those of actual humans conversing with AI (red dotted line). Humans interacting with AI also produces different conversational statistics than human-to-human conversations, highlighting that the prompt engineered base tutor model (GPT-4) impersonates a human tutor with limited fidelity. These results provide further evidence that simulated dialogues involving a fine-tuned student model approximate human-AI conversations better than conversations generated from two prompt-engineered AI models alone.  Error bars and intervals represent $95\%$ confidence intervals calculated with the Student’s t-distribution (light green and light red intervals show the error bars for human-to-human and human-to-GPT-4 conversations respectively).}
  \label{fig:student_model_statistics}
\end{figure} 

\subsection{Fine-Tuned Student Model}
Simulating students using generative AI is an active area of research, with prompt engineering being the prominent method of creating different student personas \cite{tack2022aiteachertest, wu2025embracingimperfection, li2025metacognitivecultivation, mahajan2024alignsim, hu2025LLMteachingplans}. We generally expect students to be easier to simulate than teacher models \cite{markel2023gpteach,tack2022aiteachertest}, due to the fact that in most educational dialogues students are expected to {\it follow instructions} (similar to LLMs) as opposed to {\it leading with instructions}.
However, prompt-engineered student simulators often lack the authenticity and variety that is needed to capture a wide range of different learning personas~\cite{markel2023gpteach}. 

\red{In our approach, we use student data to train a student model through parameter-efficient fine-tuning (PEFT)~\cite{hu2021loralowrankadaptationlarge,schulman2025lora}. This approach was inspired by the empirical realization that even human testers with relevant domain-expertise (e.g. the authors of this paper) are low-fidelity impersonators of students. At the same time, with access to extensive dialogue data about how students talk, behave and
learn in actual sessions it is possible to train representative student models that mimic human
learners with high fidelity, including the conversational benchmarks outlined in Section~\ref{section:proxies}.}

The steps of training a fine-tuned student model are as follows:
\begin{enumerate}
    \item \red{We select a large, random sample of projects. We assign non-descript ID's to track each project during the training process.}
    \item \red{We perform multiple epochs of supervised fine-tuning on the student turns in these transcripts with a system prompt that contains the project ID, stopping the training before overfitting begins. This training on a large amount of student data gives the model a general sense of how students typically communicate and behave, while also ingraining individual student behavior into the model parameters.}
    \item \red{We randomly choose a small subset of students ($n=10$) for actual simulations and selectively activate them by using exactly the same prompt --including the associated project-specific ID-- that was used during the fine-tuning phase.}
    \item \red{In addition, we use an LLM to extract a comprehensive set of details that the student reveals about themselves during the student-tutor conversation and include it in the system prompt. This further reinforces the fine-tuned behavior of the student. }
    \item We run single-shot evaluations on the student model to confirm that it responds to a set of selected queries as expected.
\end{enumerate}

Upon completing training, we carry out qualitative and quantitative assessments to verify that the model’s behavior closely reflects that of the original students. \red{Qualitative checks performed by Polygence personnel confirm that these fine-tuned and prompt-reinforced student models respond to questions about their academic background, project interests, etc. in a similar style as their live counterparts.}

A summary of our quantitative checks is provided in Fig.~\ref{fig:student_model_statistics}. Specifically, in Fig.~\ref{fig:student_model_statistics} we study the conversational statistics that emerge when this fine-tuned student model is paired with an off-the-shelf model (GPT-4) and compare the results in three other scenarios: (1) human-to-human interactions, (2) human-to-AI interactions, and (3) AI-to-AI interactions without fine-tuning. We specifically picked GPT-4 for this analysis because it allows us to use the conversational human-to-AI data from our PolyPilot experiment (see Section~\ref{section:polypilot}) for benchmarking.

Crucially, we find that simulated conversations between two AI base models (\textbf{Base Student Model to GPT-4}, large orange dot) produce markedly different conversational statistics from both dialogues involving only humans (\textbf{Human to Human}, green dashed line) and those between humans and an AI tutor (\textbf{Human to GPT-4}, red dotted line). A one-way Analysis of Variance (ANOVA) followed by Tukey's HSD post-hoc tests confirmed that these differences are highly statistically significant.

When interacting with GPT-4, the student base model talks less overall than a human student with a human tutor (mean of 22.2\% vs. 29.9\%, a difference of $-7.7$ percentage points, $p < 0.001$). At the same time, it talks substantially more than a human student interacting with GPT-4 (22.2\% vs. 10.5\%, a difference of $+11.8$ p.p., $p < 0.001$). Furthermore, the base model also includes significantly more questions per turn (mean of 1.91) than both the Human-to-AI scenario ($+1.47$ questions, $p < 0.001$) and the Human-to-Human scenario ($+0.80$ questions, $p < 0.001$). Finally, the base model's turns are substantially longer, averaging 66.1 words, which is significantly more than both the Human-to-AI scenario ($+40.5$ words, $p < 0.001$) and the Human-to-Human scenario ($+42.6$ words, $p < 0.001$). Notably, while the base model is a clear outlier on this metric, the average turn lengths for the Human-to-AI (25.5 words) and Human-to-Human (23.4 words) conditions were not statistically different from one another ($p = 0.589$).

In contrast, after training on our student data, the fine-tuned student model displays conversational statistics that begin to mimic those of humans interacting with AI. These results indicate that simulated dialogues involving a fine-tuned student model provide a more accurate representation of human-AI conversations than those generated from two prompt-engineered AI models alone.

\subsection{Multi-Turn Evaluations}
\label{section:multi_turn_evals}

Our student simulator allows us to run multi-turn evaluations at scale on any LLMs. First, we pick a single fine-tuned student model by selecting a training checkpoint that performs well on our evaluations and that is well aligned with human-to-AI conversational statistics (Fig.~\ref{fig:student_model_statistics}). We then initiate a conversation between the student model and a tutor model that we are looking to evaluate by feeding the output of the tutor model into the student model and vice versa. This continues until either the end of the conversation is detected by a dedicated LLM judge (Gemini 2.5 Flash), or the number of turns hit a predetermined limit (this ensures that conversations do not continue indefinitely). Once the conversation is finished, it is analyzed for conversational statistics and passed on to a reasoning LLM judge for a single-shot evaluation (see e.g. `Uncovering student background \& learning context' in Section~\ref{section:proxies}). We typically repeat this 10 times for each of the 10 student models, thereby ensuring that each evaluation point corresponds to 100 simulated multi-turn conversations. We chose these numbers to limit variability and to ensure reproducibility between consecutive evaluation experiments. 

\section{Benchmarking LLMs Against Human Performance}
\label{section:llm_benchmarking}

Before fine-tuning LLMs with authentic learning data, we first sought to evaluate the performance gap between state-of-the-art LLMs and humans in terms of conversational and pedagogical capabilities.  To establish human performance for the six selected benchmarks outlined in the previous section, we analyze 80,000 hours of data from Polygence. We also evaluate off-the-shelf LLMs for the same benchmarks. Our methodology for performing these evaluations using a fine-tuned student model is discussed in Section~\ref{section:multi_turn_evals}.

\begin{figure}[h]
  \centering
  \includegraphics[width=0.95\textwidth]{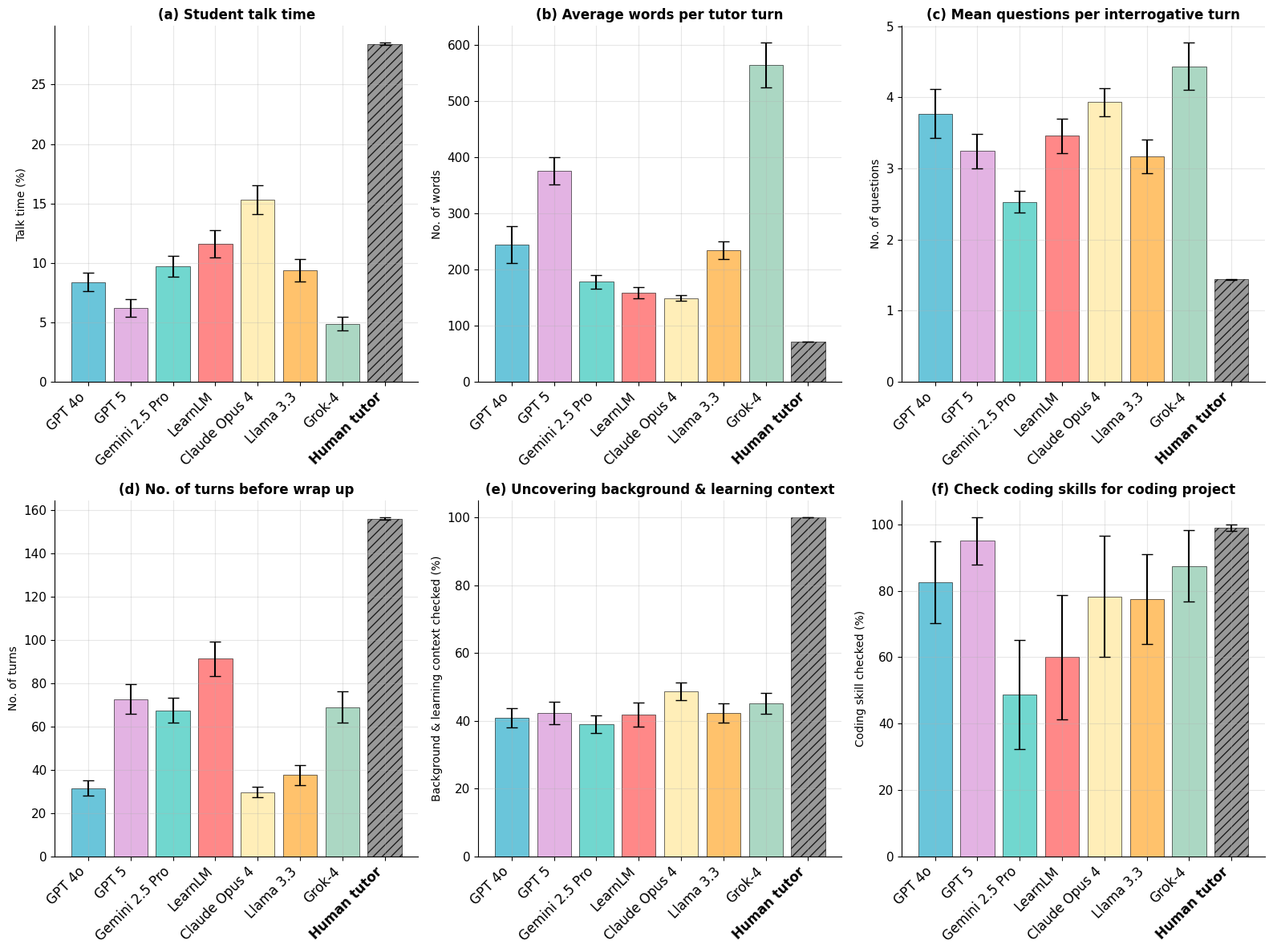}
\caption{Benchmarking results for six conversational and pedagogical metrics comparing human tutors (hatched gray) with state-of-the-art LLMs from OpenAI, Google, Anthropic, Meta, and XAI. Benchmarks include (a) student talk time, (b) average words per tutor turn, (c) mean questions per interrogative turn, (d) number of turns before wrap-up, (e) uncovering student background and learning context, and (f) checking coding skills for coding projects. Human tutors consistently outperform LLMs across these benchmarks.}

  \label{fig:human_benchmarks}
\end{figure}

Fig.~\ref{fig:human_benchmarks} shows our benchmarking results for humans (hatched gray) and a range of different state-of-the-art LLMs from OpenAI, Google, Anthropic, Meta, and XAI. In Fig.~\ref{fig:human_benchmarks}(a), we find that in human-to-human tutor-led dialogues students speak close to 30\% of the time. Off-the-shelf models are typically more verbose than humans, leaving only 5-15\% of talk time to students. A hallmark of dynamic dialogues is when students and tutors take frequent turns, rather than one party monopolizing the conversation. Fig.~\ref{fig:human_benchmarks}(b) shows that on average, human tutors speak only 72 words before passing the turn to the student. In contrast, most off-the-shelf models average 150-300 words per turn. Another important technique in human tutoring is asking open-ended questions and then pausing and letting students answer. Statistically speaking, human tutors typically ask 1-2 questions per interrogative turn (on average 1.5 in our dataset). In Fig.~\ref{fig:human_benchmarks}(c) we observe the inherent tendency of LLMs to ask a high number of questions (3-4 per interrogative turn). Another shortcoming of LLMs is the rapid drive towards `resolving' conversations. Fig.~\ref{fig:human_benchmarks}(d) shows that most LLMs tend to end student conversations in 30-80 turns, which contrasts with human tutors, whose average session length is closer to 150-160 turns (note that session lengths are widely distributed, with a mean of 56 minutes and a standard deviation of 16 minutes despite the nominal 1-hour length).

Fig.~\ref{fig:human_benchmarks}(e) shows our most complex evaluation, which quantifies the extent to which off-the-shelf frontier models are able to uncover relevant information about the student's background and learning context. We find that most models score in the 40-45\% interval, often missing key context (such as coding skills, goals, needs, motivation, etc.) about the student.

Fig.~\ref{fig:human_benchmarks}(f) shows results for a more specific version of the previous benchmark -- the tendency of models to check the student's the coding background in cases where their project requires coding (a binary yes/no decision). Human data shows that tutors check students’ coding background virtually any time the project
involves coding. In contrast, we find that most off-the-shelf models perform in the 50-80\% range,
with only a few thinking models doing better.

Finally, we note that these benchmarking experiments have generally proven reproducible across runs and across different versions of our student model. For example, we consistently find that among the models we analyzed, Claude produces the highest student talk time, LearnLM maintains the conversation the longest, and GPT-5 (and previously o1) performs the best on checking the coding skills of the student.

\section{Fine-Tuning on Authentic Learning Data}
\label{section:model_tuning}

\begin{figure}[h!]
    \centering



    \begin{subfigure}[b]{0.32\textwidth}
        \centering
        \includegraphics[width=\linewidth]{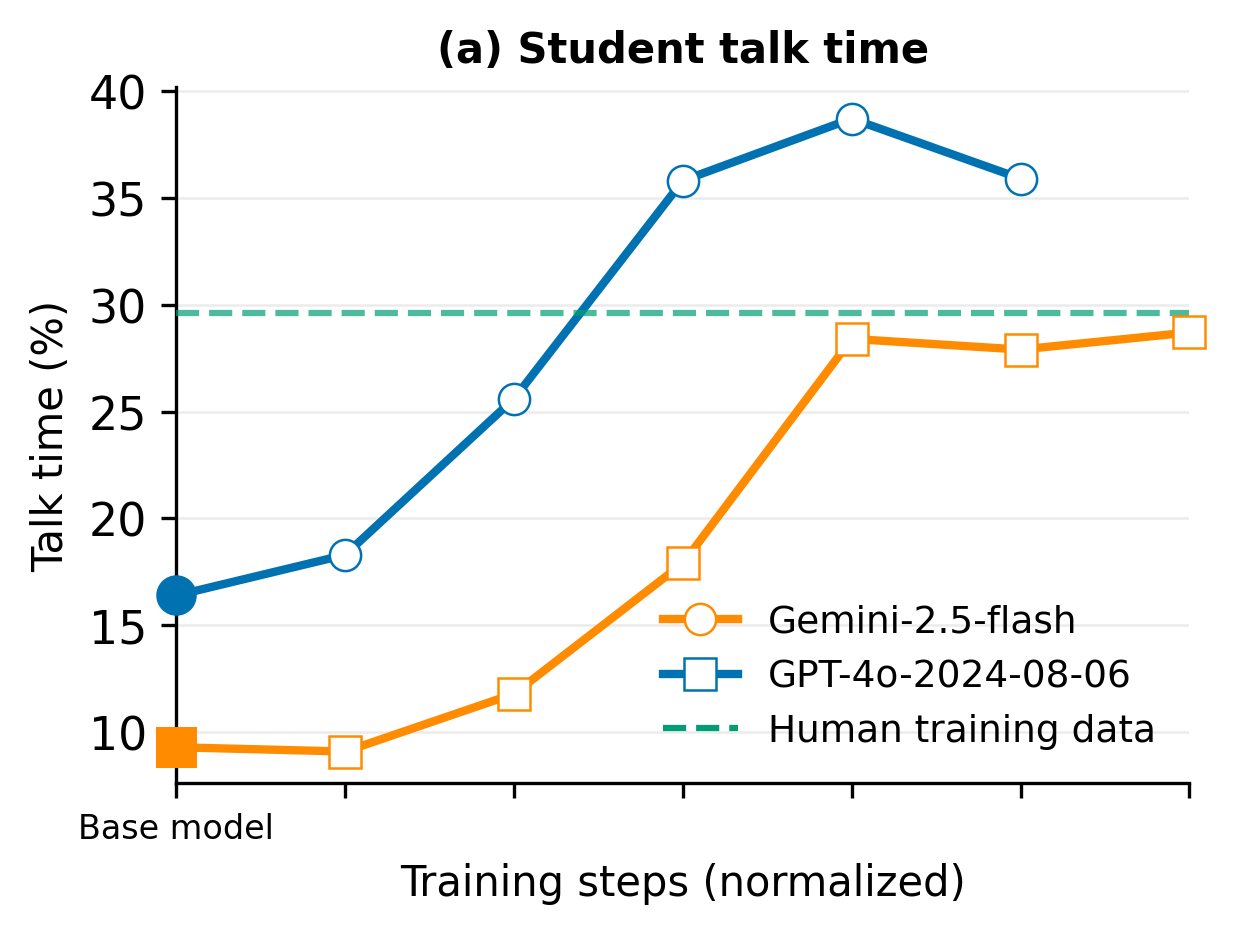}
        \label{fig:talk_time}
    \end{subfigure}\hfill    
    \begin{subfigure}[b]{0.32\textwidth}
        \centering
        \includegraphics[width=\linewidth]{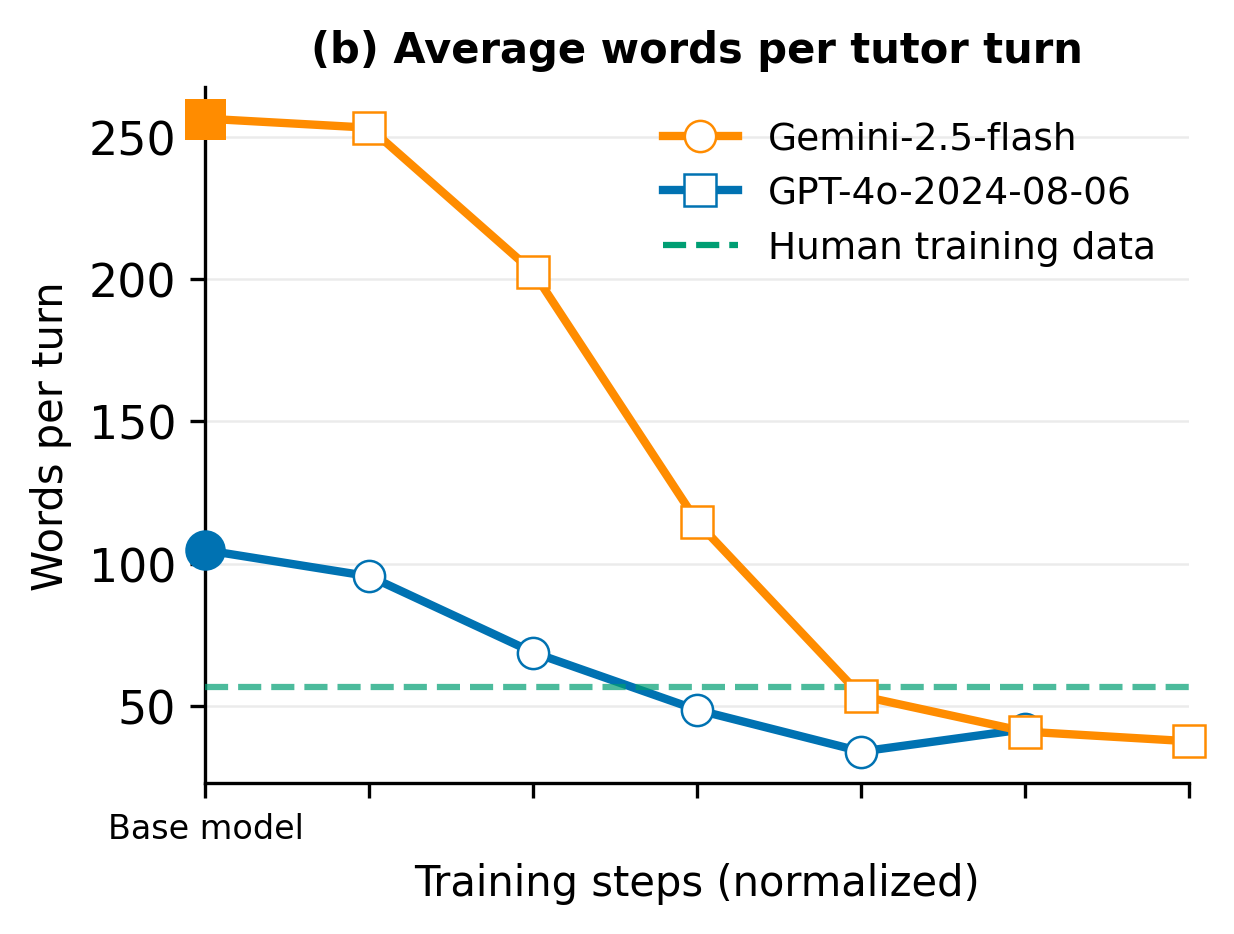}
        \label{fig:avg_words}
    \end{subfigure}\hfill
    \begin{subfigure}[b]{0.32\textwidth}
        \centering
        \includegraphics[width=\linewidth]{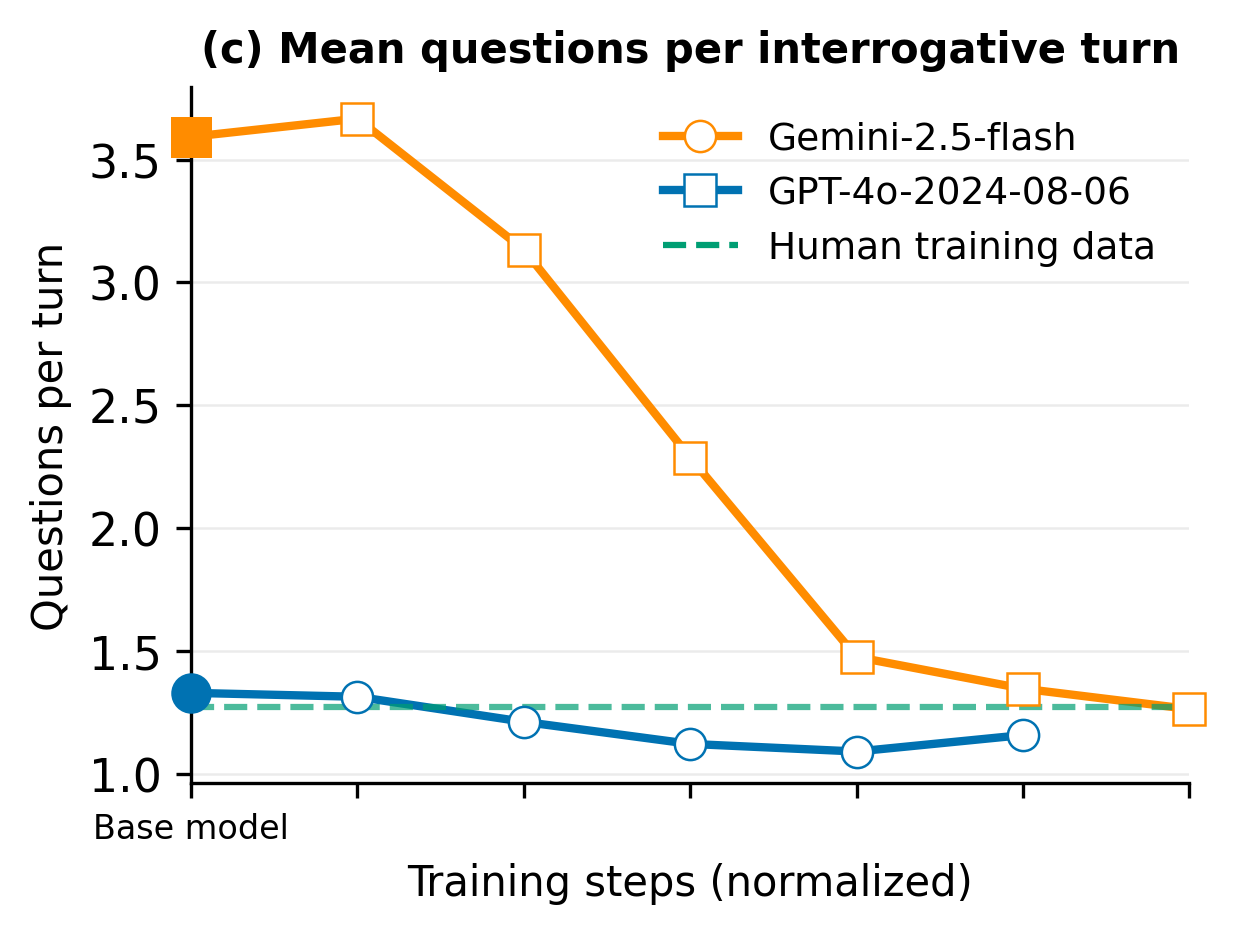}
        \label{fig:question_ratio}
    \end{subfigure}

    \vspace{1em}

    \begin{subfigure}[b]{0.32\textwidth}
        \centering
        \includegraphics[width=\linewidth]{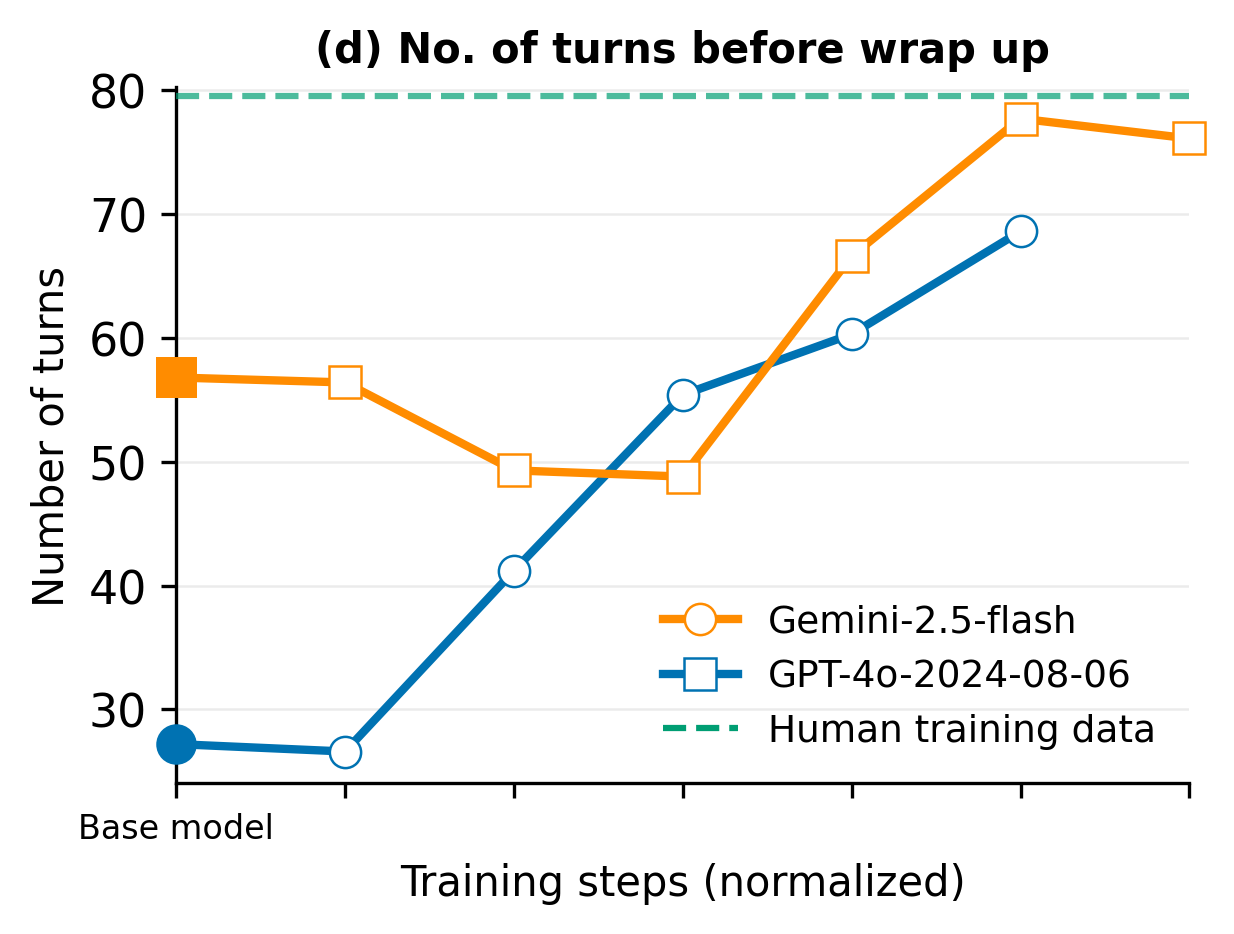}
        \label{fig:num_turns}
    \end{subfigure}\hfill    
    \begin{subfigure}[b]{0.32\textwidth}
        \centering
        \includegraphics[width=\linewidth]{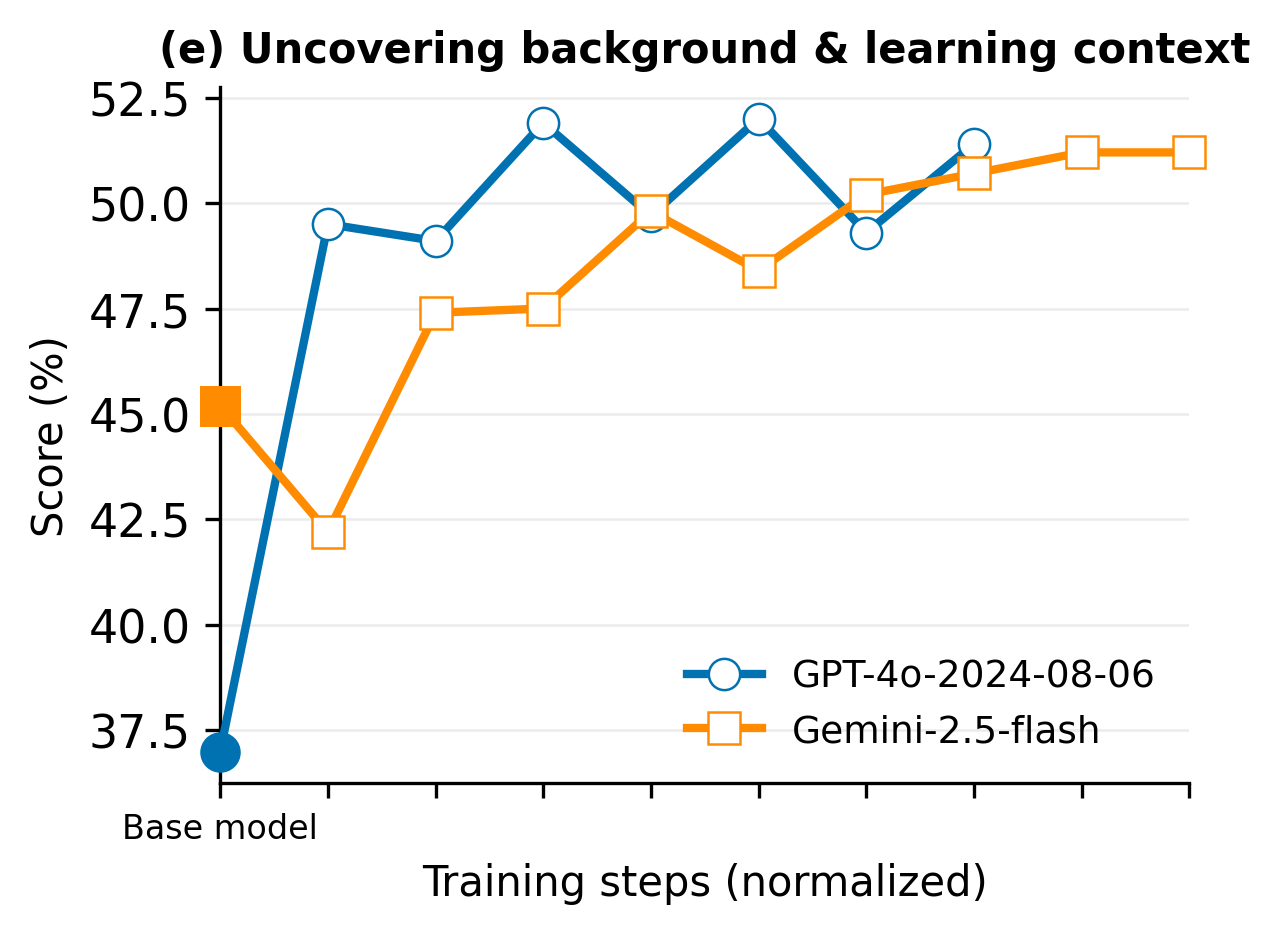}
        \label{fig:student_info_checklist}
    \end{subfigure}\hfill  
    \begin{subfigure}[b]{0.32\textwidth}
        \centering
        \includegraphics[width=\linewidth]{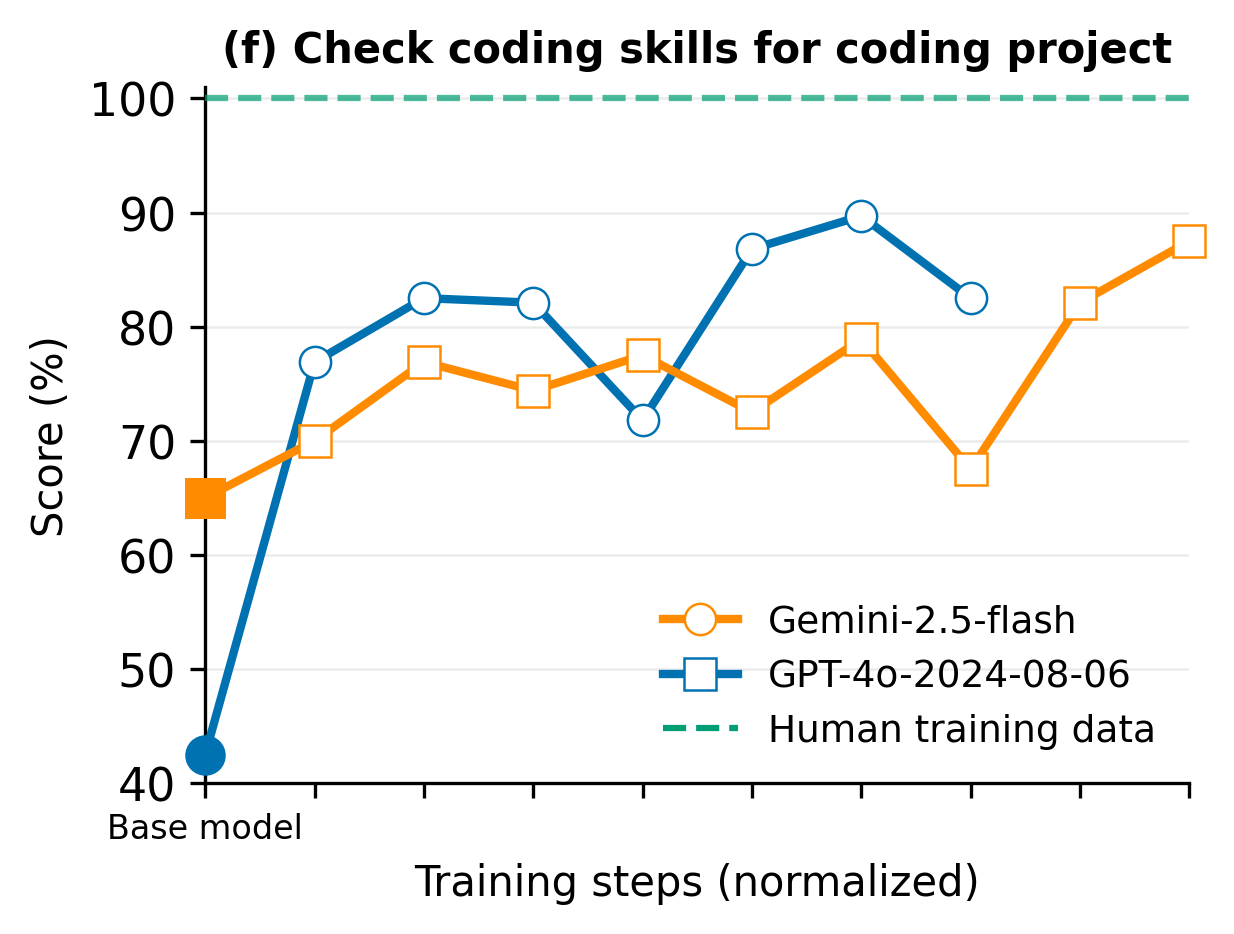}
        \label{fig:coding_background}
    \end{subfigure}

\caption{Fine-tuning results on six conversational and pedagogical benchmarks for Gemini 2.5 Flash (orange) and GPT-4o-2024-08-06 (blue). The dashed green line indicates performance from human training data. Benchmarks include (a) student talk time, (b) average words per tutor turn, (c) mean questions per interrogative turn, (d) number of turns before wrap-up, (e) uncovering student background and learning context, and (f) checking coding skills for coding projects. Fine-tuning consistently improves model performance, with simpler conversational metrics (a–d) converging more quickly than the more complex pedagogical benchmarks (e–f).}

\label{fig:mentor_metrics}
\end{figure}

\red{In this section, we demonstrate that performing parameter-efficient fine-tuning~\cite{hu2021loralowrankadaptationlarge,schulman2025lora} on state-of-the-art frontier models using high-quality post-training data (see Section~\ref{section:polygence_data}) improves their performance on the benchmarks outlined in Section~\ref{section:proxies}.} Specifically, we fine-tune Google's Gemini 2.5 Flash and OpenAI's GPT 4o-08-06, both of which are state-of-the-art models and are the most recent non-thinking model versions\footnote{For Gemini 2.5 Flash we \href{https://ai.google.dev/gemini-api/docs/thinking}{set the thinking budget} to zero tokens.} from their respective providers with API access to fine-tuning. 

\subsection{Benchmarking through automatic evaluations} 
Fig.~\ref{fig:evals_for_selected_checkpoint} shows a representative set of results of our fine-tuning experiments across the six different benchmarks under consideration. The dashed green line shows results from the human training data for reference (note that these results are slightly different from Fig.~\ref{fig:human_benchmarks} due to the effects of the cleaning process outlined in Section~\ref{section:transcript_processing}). We see that fine-tuning improves model performance on every benchmark, bringing the resulting values closer to the human training data benchmarks. Specifically, we find that student talk time increases, the average number of words per tutor turn goes down, the mean number of question per interrogative turn settles in the 1-2 question region\footnote{Note that GPT-4o-2024-0806 ask comparatively fewer questions (around 1.3) per interrogative turn than other off-the-shelf models. Nonetheless, fine-tuning changes how it is formulating its questions. We found that later models from OpenAI -- such as GPT-4o-2024-11-20, o1, and GPT-5-- regress to asking 3-4 questions per interrogative turn.}, and the number of conversation turns increases. We also find that models show marked improvements on the complex evaluation of uncovering student background and learning context, although the absolute values are still far from 100\%. We also find a corresponding improvement on the coding skill check benchmark. 

The $x$-axes in Figs.~~\ref{fig:evals_for_selected_checkpoint}(a)-(d) represent a smaller number of training steps than Figs.~~\ref{fig:evals_for_selected_checkpoint}(e)-(f). This aligns with our empirical observation that training models to uncover student background, learning context, and coding skills requires significantly more time than training them on simpler conversational improvements. Finally, the near-monotonic improvement on all benchmarks as training progresses allows the selection of checkpoints, which show clear improvement on all benchmarks simultaneously (see Appendix~\ref{app:tuned_checkpoint} for an example). 

\subsection{Human Evaluations}
In addition to the large-scale automated evaluations outlined above, we also performed limited human evaluations of the resulting fine-tuned models to confirm the reported performance improvements. In these manual experiments, human testers (internal team members) impersonated students and entered into dialogues with the fine-tuned tutor model. This allowed us to directly observe reduced verbosity and enhanced turn taking, a marked shift towards fewer but more open-ended questions, a more natural flow of conversations, and clear improvement in context-setting and in-depth understanding of student background before diving into further tutoring activity. A rigorous, large-scale human evaluation of our fine-tuned models will be addressed in an upcoming report.

\section{Conclusion}

In this technical report, we analyzed the fundamental limitations of prompt-engineering LLMs for education and outlined the importance of post-training frontier models on authentic learning data. We introduced a novel framework for performing multi-turn evaluations on LLMs using a fine-tuned student model trained on authentic student data. We used our framework to quantify the gap between frontier models and human tutors in terms of six conversational and pedagogical benchmarks. We also showed that fine-tuning on authentic tutor data improved the performance of frontier models on all of our benchmarks.

\section{Outlook}

This preliminary report focused on the supervised fine-tuning of frontier models, which is only the first step in post-training of LLMs. Encouraged by these early results, we are now focused on realizing the full benefit of post-training through reinforcement learning from human feedback (RLHF) \cite{ziegler2019finetuning}, which is a natural next step given the availability of authentic learning dialogues involving humans. 

Our early efforts focused on establishing the simple, measurable benchmarks outlined in this report, but more sophisticated evaluations are needed to capture the full richness of human pedagogy. Specifically, we will prioritize creating benchmarks that capture the nuances of longitudinal interactions between students and tutors. These types of interactions are only possible over extended periods of time and are critical in fostering rapport and driving tangible learning outcomes. 

We also note that evaluations in this report were focused on scalable and automated processes and (limited) human feedback from non-learners. Our aim is to scale and quantify {\it student feedback} on actual model performance by incorporating post-trained models into the student journey on the Polygence platform.

\section{Acknowledgements}
We acknowledge helpful discussions with Hossein Talebi, Mike Tung, Rumen Dangovski, Peter Danenberg, Tony Wang, Kelvin Guu, James Kim, Ann Miura-Ko, Vinit Sukhija, Chris Piech, Irina Jurenka, Hema Baja, Lisa Wang, Markus Kunesch, Veronica Edwards, Adam Gyulavari, and Daniel Horvath.

The PolyPilot experiment was created by Robert Beretka, Daniel Horvath, Ross Greer, Janos Perczel, Emma Leyden, Amanda Chagoya, Alex Armstrong, Abdelaziz Tina, Tamas Csaba Kadar. 

\section*{Author Information}
Dr. Demszky's contribution to this publication was as a consultant and was not part of her Stanford University duties and responsibilities. 
Janos Perczel is the co-founder and CEO of Polygence, and Jin Chow is the co-founder and COO of Polygence. 

\section*{Author Contributions}
J.P. conceived the project, developed the methodology, and executed the technical work. 
D.D. contributed technical advice and ideas. 
J.C. assisted with data cleaning and model testing. 
J.P., J.C. and D.D. wrote the paper.

\bibliographystyle{unsrtnat}
\bibliography{refs}

\appendix
\section{PolyPilot Experiment: Building a Prompt-Engineered Tutor}
\label{app:polypilot}

\begin{figure}[h]
  \centering
  \includegraphics[width=0.95\textwidth]{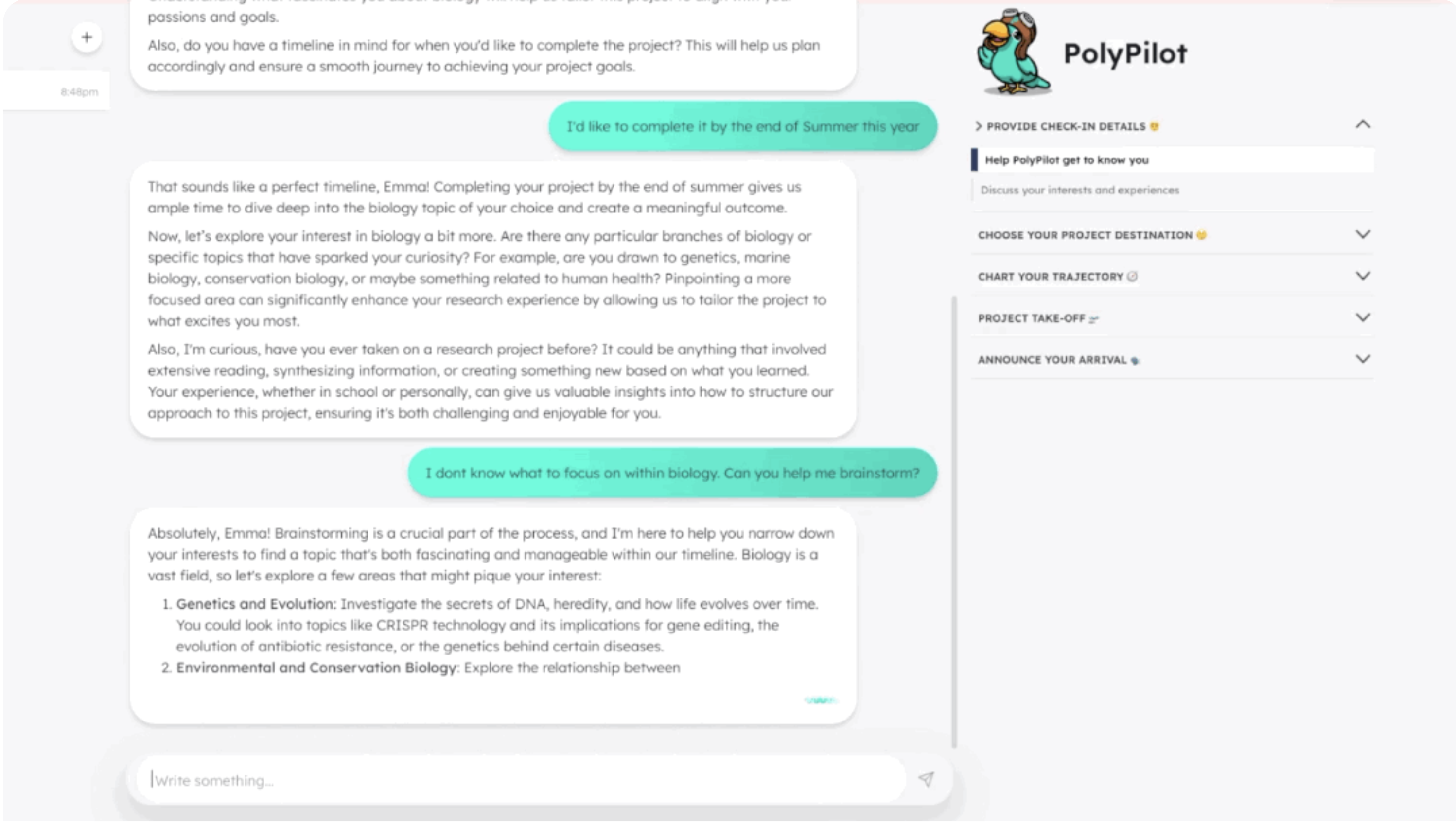}
\caption{Screenshot of the PolyPilot interface. The right-hand side shows the different pre-defined stages in the student-journey}

  \label{fig:polypilot_screenshot}
\end{figure}

PolyPilot was a high-conviction product bet in early 2024 to build an effective AI tutor for project-based learning (``research mentor'') by prompt engineering GPT-4. PolyPilot was designed to be a ChatGPT-like interface, but adapted to the needs of students looking to complete a long-term project (see screenshot in Fig.~\ref{fig:polypilot_screenshot}). The student journey was broken up into five distinct stages:
\begin{itemize}
    \item {\bf Provide check-in details} -- This stage focused on establishing context about the student's background, interests, and goals.
    \item {\bf Choose your project destination} --  This stage helped students scope out their project idea.
    \item {\bf Chart your trajectory} -- This stage guided the students through the background work needed to complete the project. 
    \item {\bf Write your first draft} -- This stage focused on starting the writing process.
    \item {\bf Announce your arrival} -- This stage helped students finish writing their artifact and showcasing it to the public.
\end{itemize}

For each stage we customized our prompts and upon detecting stage completion, the prompt for the next session was loaded on the backend. We iteratively refined the prompts to better align GPT-4 with good pedagogy and to handle edge cases. Initially, we experimented with just five prompts for the five distinct stages and the prompts quickly ballooned to 500-1000 words in length. We observed severe limitations in GPT-4's ability to follow all instructions. Later, we broke down each of the 4 stages into 3-6 distinct sub-stages and progressively refined the prompts to be only 200-300 words. These improved the performance, but the overall pedagogical capabilities of the product remained limited. 
To improve product quality and reliability, we introduced human checkpoints where an actual tutor checked the progress of the students and decided whether the student would need to do more work before proceeding to the next stage. We found that these human checkpoints were helpful, but didn't fully address the underlying issues. The product was used by $n=71$ students generating valuable data about student-AI interactions. 

\section{Observations on Anthropic's \emph{Learning Mode}, OpenAI's \emph{Study Mode}, and Google's \emph{Guided Learning}}
\label{app:study_modes}

In recent months, multiple model developers have released educational LLMs that have been prompt-engineered to improve their pedagogical behavior. To get a directional sense of their behavior, we manually tested Anthropic's \emph{Learning Mode}~\cite{anthropic2024claudeedu}, OpenAI's \emph{Study Mode}~\cite{openai2024studymode}, and Google's \emph{Guided Learning}~\cite{google2024guided} (which integrates LearnLM~\cite{modi2025evaluating}). Topics ranged from solving simple quadratic equations, to learning Latin, to studying the advanced physics of black holes. Goals ranged from learning new topics, to refining existing knowledge, to writing research papers. Among others, we found the following patterns:
\begin{itemize}
    \item {\bf Missing learning context}: All three models spend practically no time establishing the learning context, such as the pre-existing level of understanding, the learning goal or the motivation of the student. As a result, models dive quickly into discussions of specific topics that may be inappropriate for the level of the student or drive towards outcomes that do not reflect the learning goal of the student. In cases where the model asked questions about the student's existing level of understanding, we found limited evidence that this was taken into account during the conversation. 
    \item {\bf Multiple-choice-style questioning:} All three models appear to struggle with asking open-ended questions, instead defaulting to multiple-choice style questions (typically 3) that dramatically restrict the space of choices. We theorize that this is a consequence of the underlying off-the-shelf model behavior that is learned from data containing bullet-point-style, overly-structured responses. This typically narrows the topics and/or course of action quickly without appropriately exploring the ideas and goals of the student.
    \item {\bf One question per statement:} We found that Gemini and GPT typically ask exactly one question per statement (occasionally two), usually placed at the end of the statement, which we theorize is a simple consequence of a prompt instruction. While this behavior is an improvement over the large number of questions off-the-shelf models typically ask (see Section~\ref{section:llm_benchmarking}), this simple, static rule represents a coarse approximation of high-quality Socratic questioning. (We found that Claude asks a large number of questions per statement similar to its off-the-shelf version.)
    \item {\bf Wall of text:} All three models are verbose (though less so than their off-the-shelf versions), which appears to get worse as the conversation progresses. We theorize that the underlying training data ingrains verbosity into models and the effectiveness of steering models towards brevity with prompts diminishes as a progressively larger part of the context window is taken up by the conversation. 
    \item {\bf Inability to deal with confusion:} Not giving away answers has been a prominent (and potentially over-simplified) focal point of improving LLMs for education. GPT appears to still give away answers with minimal prompting. Gemini and Claude immediately and repeatedly rephrase and reexplain the question or problem when faced with confusion -- making little effort to understand the source or level of the confusion. 
\end{itemize}

We note that while each of the specific behaviors highlighted above could be improved through targeted prompting, these issues are merely indicators of a larger issue with prompt engineering. As noted previously (and highlighted in the LearnLM team's initial report~\cite{learnlm2024}), prompt engineering alone is unlikely to fully encode the vast and complex space of effective tutoring, as it would require an exhaustive, context-specific rule-based description of all of good pedagogy -- an impossible task even with unlimited context length.

\section{Overlap of Polygence Data with Top Student Activities Reported by OpenAI}
\label{app:openai_overlap}

We map the tutoring activities in our dataset to the top 27 categories of ChatGPT usage by 18-24 year old students, as reported by OpenAI \cite{openai2025collegechatgpt}. We find that approximately 78\% of our data overlaps with the top 10 ChatGPT use cases by students. Table~\ref{table:overlap_table} below shows the specific percentage of students utilizing each of the top 10 activities on OpenAI's platform. We separately show whether that activity is covered by our data.

\begin{table}[h!]
\centering
\begin{tabular}{|l|c|c|}
\hline
\makecell{\textbf{Top 10 ChatGPT Usage}\\\textbf{Categories by Students}} & 
\textbf{\% of Students} & 
\textbf{Data Overlap} \\ \hline
Starting papers/projects & 49\% & \cmark \\ \hline
Summarize texts & 48\% & \xmark \\ \hline
Brainstorm creative projects & 45\% & \cmark \\ \hline
Explore topics & 42\% & \cmark \\ \hline
Edit writing & 42\% & \cmark \\ \hline
Mathematical problem-solving & 39\% & \cmark \\ \hline
Exam preparation & 36\% & \xmark \\ \hline
Academic research & 34\% & \cmark \\ \hline
Tutoring & 32\% & \cmark \\ \hline
Essay drafting & 32\% & \cmark \\ \hline
\end{tabular}
\caption{Overlap of Polygence data with the top 10 ChatGPT use cases by 18-24 year old students, as reported by OpenAI \cite{openai2025collegechatgpt}.}
\label{table:overlap_table}
\end{table}

\section{Selecting a Checkpoint With Improved Performance Across All Benchmarks}
\label{app:tuned_checkpoint}

\begin{figure}[h]

  \centering
  \includegraphics[width=0.85\textwidth]{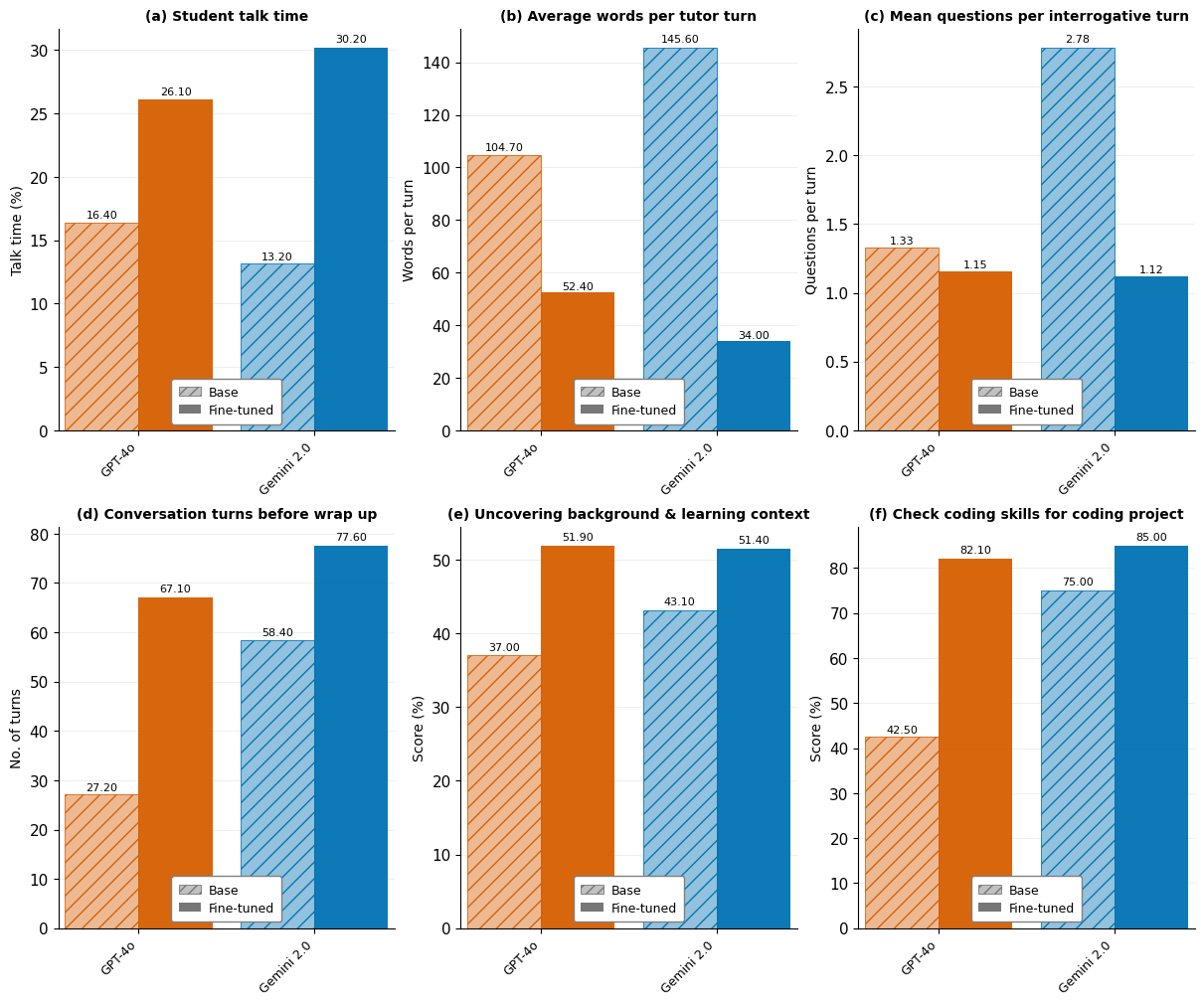}
  \caption{Performance of a specific checkpoint a fine-tuned Gemini 2.0 model on all six benchmarks from Section~\ref{section:proxies}. We observe simultaneous improvement across all benchmarks.}
  \label{fig:evals_for_selected_checkpoint}
\end{figure}

In Section~\ref{section:model_tuning} we showed that models show general improvement across all selected benchmarks when trained on authentic learning data. In Fig.~\ref{fig:evals_for_selected_checkpoint} we choose a specific checkpoint and show that the fine-tuned version outperforms the base model on all six of the benchmarks simultaneously.

\end{document}